\documentclass{article}

\usepackage{microtype}
\usepackage{graphicx}
\usepackage{subcaption}
\usepackage{booktabs} % for professional tables

\usepackage{hyperref}

\newcommand{\vcenteredinclude}[1]{\ensuremath{\vcenter{\hbox{\includegraphics[width=\linewidth]{#1}}}}}
\usepackage[preprint]{icml2026}

\usepackage{amsmath}
\usepackage{amssymb}
\usepackage{mathtools}
\usepackage{amsthm}

\usepackage[capitalize,noabbrev]{cleveref}

\theoremstyle{plain}

\theoremstyle{definition}

\theoremstyle{remark}

\usepackage[textsize=tiny]{todonotes}
\icmltitlerunning{Structural Disentanglement in Bilinear MLPs via Architectural Inductive Bias}

\usepackage{xcolor}

\definecolor{BilinearBlue}{HTML}{0000FF}
\definecolor{ReluRed}{HTML}{FF0000}

\begin{document}

\twocolumn[
  \icmltitle{Structural Disentanglement in Bilinear MLPs via Architectural Inductive Bias}



\icmlsetsymbol{equal}{*}

  \begin{icmlauthorlist}
    \icmlauthor{Ojasva Nema}{equal,comp}
    \icmlauthor{Kaustubh Sharma}{equal,yyy}
    \icmlauthor{Aditya Chauhan}{equal,xxx}
    \icmlauthor{Parikshit Pareek}{yyy}
  \end{icmlauthorlist}

    % \begin{icmlauthorlist}
    % \icmlauthor{Firstname1 Lastname1}{}
  % \end{icmlauthorlist}

  \icmlaffiliation{yyy}{Department of Electrical Engineering, Indian Institute of Technology Roorkee (IIT Roorkee), India}
  \icmlaffiliation{comp}{Department of Material Science \& Engineering, IIT Roorkee, India}
   \icmlaffiliation{xxx}{Mehta Family School of Data Science and Artificial Intelligence, IIT Roorkee, India}

  \icmlcorrespondingauthor{Parikshit Pareek}{pareek@ee.iitr.ac.in}
  \icmlcorrespondingauthor{This work was supported by the ANRF PM Early Career Research Grant (ANRF/ECRG/2024/001962/ENS), the IIT Roorkee Faculty Initiation Grant (IITR/SRIC/1431/FIG-101078).}{https://psquare-lab.github.io}

  % % will not be shown in the document
  \icmlkeywords{Machine Learning, ICML}

  \vskip 0.3in
]



\printAffiliationsAndNotice{\icmlEqualContribution}   

\begin{abstract}
Selective unlearning and long-horizon extrapolation remain fragile in modern neural networks, even when tasks have underlying algebraic structure. In this work, we argue that these failures arise not solely from optimization or unlearning algorithms, but from how models structure their internal representations during training. We explore if having explicit multiplicative interactions as an architectural inductive bias helps in structural disentanglement, through Bilinear MLPs. We show analytically that bilinear parameterizations possess a `non-mixing' property under gradient flow conditions, where functional components separate into orthogonal subspace representations. This provides a mathematical foundation for surgical model modification. We validate this hypothesis through a series of controlled experiments spanning modular arithmetic, cyclic reasoning, Lie group dynamics, and targeted unlearning benchmarks. Unlike pointwise nonlinear networks, multiplicative architectures are able to recover true operators aligned with the underlying algebraic structure. Our results suggest that model editability and generalization are constrained by representational structure, and that architectural inductive bias plays a central role in enabling reliable unlearning.
\end{abstract}

\section{Introduction}

Neural networks have become the dominant paradigm for learning complex function classes across domains ranging from scientific computing to vision and language. While there has been progress that has focused on increasing scale and expressivity, there is growing recognition that performance alone is not sufficient. Two additional capabilities that are of increasing importance are: (1) \textit{Generalization}, the ability to extrapolate to solve a derived task not directly in the training data, and (2) \textit{Unlearning}, the ability to selectively modify model behavior after training. Even when training performance is near-perfect, modern models often perform poorly on these fronts.

Recent work \cite{pearce2025bilinear, power2022grokkinggeneralizationoverfittingsmall,nanda2023progress,gromov2023grokkingmodulararithmetic} suggests that these properties are closely related to the structural consistency with which a model represents the task. Networks that internalize the algebraic structure of a problem tend to exhibit improved interpretability, extrapolation, and unlearnability. In contrast, networks that rely on surface-level approximation often fail to do so, despite achieving similar training accuracy, showing importance of how a task is internally represented.

Despite this, most existing approaches to unlearning treat it as a procedural problem to be solved post-hoc. \cite{kurmanji2023unboundedmachineunlearning,hong2024dissectingfinetuningunlearninglarge,10.1145/3749987}. While these methods can be effective in certain cases, they implicitly assume that the underlying representations are, in principle, conceptually separable. When representations are fundamentally entangled, and the parameters jointly encode multiple behaviors, no algorithm can reliably isolate capabilities. This raises a more basic question: \textit{When is selective unlearning structurally possible?}

In this work, we argue that this question must be answered at the level of representation, and thus, architecture. We introduce the notion of structural disentanglement: a property of learned representations in which the model's internal operator decomposes into orthogonal, independent components aligned with the task's underlying compositional structure. We formulate this notion at the operator level, distinct from neuron-level modularity. We assert that if distinct functional components evolve orthogonally during training, they can be modified independently, rather than focusing on modular neurons or post-hoc interpretability methods. Moreover, architectural inductive bias plays a key role in determining whether such structure can emerge in a class of problems, since it constrains the geometry of the function class itself. 

Piecewise-linear activations such as ReLU have proven highly effective in practice, offering strong approximation properties and stable optimization while training. Their additive and piecewise structure makes the network represent functions as sums of locally linear regions. However, many tasks of interest are governed by multiplicative or compositional rules. Examples include algebraic operations such as modular arithmetic, cyclic reasoning updates, Lie group dynamics in physics, and multiplicative feature interactions etc. Capturing such structure requires an inductive bias that permits multiplicative interactions and respects the separability of functional components to be learned.

To study aforementioned hypothesis in a controlled and analytically tractable manner, we focus on Bilinear multilayer perceptrons \cite{7410527,li2017factorizedbilinearmodelsimage,Chrysos_2021}. Bilinear layers calculate products between learned linear projections of the input, enabling the representation of low-rank multiplicative structure. Importantly, bilinear networks remain simple, fully differentiable, and comparable in capacity to standard MLPs. Also, bilinear architectures admit a natural operator interpretation: each output can be associated with an interaction matrix that captures how input dimensions combine \cite{pearce2025bilinear}. This makes them an ideal probe for studying how architectural inductive bias shapes learned operators and its dynamics. Notably, gated activations such as GeGLU and SwiGLU also partially inherit similar biases, suggesting broader relevance.

\textbf{Positioning and Contributions:}
In this work, we focus on bilinear MLPs, motivated by theoretical insights from matrix factorization, which suggest that, in ideal gradient flow settings, explicit multiplicative parameterizations tend to preserve the independence of interaction modes during learning under standard assumptions. 

Here, we do not propose bilinear architectures as a universal replacement for pointwise nonlinear activations like ReLU or modified versions like GLU \cite{dauphin2017languagemodelinggatedconvolutional}. Rather, we use bilinear MLPs as a lens to study a specific hypothesis that: \textit{In problem regimes governed by compositional or algebraic structure, architectures that explicitly model multiplicative interactions are more likely to align with the task’s underlying independent components}. Our contributions are to \textbf{(1)} synthesize insights from gradient dynamics with the practical goals of unlearning, showing analytically that bilinear activations preserve the independence of interaction modes to provide a mechanism for precise model editing for compositional problems; \textbf{(2)} provide empirical evidence in controlled compositional/algebraic settings demonstrating that this structural alignment yields significantly higher unlearning selectivity compared to standard ReLU baseline; and \textbf{(3)} validate the generalization capability of bilinear MLPs to extrapolate to distinct algebraic structures beyond their training regimes, particularly in long-horizon compositional tasks. By reframing unlearning
as a representational property rather than an algorithmic trick, this work highlights architecture as a primary determinant of model editability and systematic generalization.

\section{Related Works}

\subsection{Machine Unlearning}
The problem of removing specific data or capabilities from trained models has been developed into the field of \emph{machine unlearning} \cite{cao2015towards}. Several surveys now provide a comprehensive overview of unlearning methods, their guarantees, and evaluation protocols \cite{bourtoule2021machine,xu2023machine_unlearning_survey,liu2025survey}.

A prominent line of work studies unlearning through \emph{parameter modification}, often via pruning or sparsification. For example, \cite{wang2022federated} use structured channel pruning to remove category-specific information in convolutional networks, while \cite{jia2023model} apply one-shot magnitude pruning to reduce the distance between an unlearned model and a fully retrained reference. These works empirically show that model sparsity can reduce unlearning error, but do not characterize when such pruning can be performed without interfering with retained functionality.

Another class of methods performs unlearning by modifying the optimization objective. \cite{trippa2024nablataugradientbasedtaskagnostic} propose gradient-based unlearning that maximizes the loss on forgotten data, while \cite{bu2025unlearningmultitaskoptimizationnormalized} formulate unlearning as a multi-task optimization problem using normalized gradient differences. Relatedly, \cite{eldan2024whos} demonstrate that selective fine-tuning can erase class-specific knowledge in large language models. While effective in practice, these approaches remain procedural: they do not explain how or why gradient updates sometimes interfere with retained tasks. In contrast to prior work, we do not introduce a new unlearning algorithm. Instead, we analyze \emph{when unlearning is structurally possible} as a consequence of the model’s parameterization and gradient dynamics.

\subsection{Bilinear and Quadratic Neural Networks}

Bilinear and quadratic neural networks introduce multiplicative interactions as an architectural inductive bias. \cite{fan2020universal} study quadratic neural networks and establish strong expressivity and universal approximation properties. More recently, bilinear-activation MLPs have been shown to admit an explicit operator interpretation, where each output class corresponds to a learned interaction matrix that can be analyzed using spectral methods \citep{pearce2025bilinear}.

While prior work on bilinear models focuses on expressivity, interpretability, or generalization, it does not analyze the optimization geometry of these architectures in the context of unlearning, nor does it characterize how gradient dynamics affect the stability of removing specific functional components.

\subsection{Structured Representations and Modularity}
Several lines of work suggest that neural networks learn modular internal representations when their architectures and training objectives align with underlying algebraic or algorithmic structure of the task at hand. In modular arithmetic tasks, for example, successful learning is associated with the emergence of interpretable, frequency-based components \citep{power2022grokking,gromov2023grokking_modular,nanda2023progress}. Similar observations have been made in arithmetic reasoning models, where architectural inductive biases influence the structure of learned representations \citep{zhou2024what,lee2024teaching}. While these works study generalization and extrapolation, they also suggest a broader principle: when learned functions decompose into modular components, model behavior may become easier to interpret.

\section{Background}\label{sec:background}

We denote scalars by lowercase italics ($s$), vectors by bold lowercase ($\mathbf{v}$), matrices by bold uppercase ($\mathbf{M}$), and tensors by sans-serif ($\mathsf{T}$). For matrix $\mathbf{M}$, $\mathbf{m}_{i:}$ is the $i$-th row. The symbol $\odot$ denotes the element-wise Hadamard product.

\textbf{Standard MLPs and Entanglement.} Standard MLPs utilize pointwise nonlinearities (e.g., ReLU) to partition the input space into convex polytopes, acting as affine transformations within each region \citep{montufar2014number}. While expressive, this piecewise-linear construction entangles representations \citep{balestriero2020mad}. Consequently, modifying weights to suppress specific behaviors risks catastrophic interference \citep{kirkpatrick2017overcoming}, and extrapolation fails because the function simplifies to linearity outside the training support \citep{xu2021how}.

\textbf{Bilinear Architectures and Interaction Operators}.
Bilinear MLPs, a simplification of Gated Linear Units (GLUs) \citep{shazeer2020gluvariantsimprovetransformer}, omit pointwise nonlinearities entirely while retaining competitive performance \citep{pearce2025bilinear}. A bilinear layer with weights $W, V \in \mathbb{R}^{d_{out} \times d_{in}}$ computes:
\begin{equation}
    g(\mathbf{x}) = (W\mathbf{x}) \odot (V\mathbf{x}).
\end{equation}
This operation can be expressed via pairwise interactions. For the $k$-th output neuron, the activation is $g_k(\mathbf{x}) = \mathbf{x}^\top (\mathbf{w}_k \mathbf{v}_k^\top) \mathbf{x}$. We define the rank-1 matrix $M_k = \mathbf{w}_k \mathbf{v}_k^\top$ as the \textit{interaction matrix} for output $k$.

To analyze a specific task or output direction defined by a vector $\boldsymbol{\alpha}$, we compute the weighted sum of these interactions, yielding the \textbf{task-specific interaction operator} $Q$:
\begin{equation}
    Q = \sum_k \alpha_k M_k = \sum_k \alpha_k \mathbf{w}_k \mathbf{v}_k^\top.
    \label{eq:operator_Q}
\end{equation}
Without loss of generality, we treat $Q$ as symmetric, as the anti-symmetric component contributes zero to the quadratic form $\mathbf{x}^\top Q \mathbf{x}$ and doesn't affect optimization dynamics. This ensures $Q$ has real eigenvalues and orthogonal eigenvectors, providing a basis for structural disentanglement.

\section{Theoretical Analysis}
\label{sec:theory}

We analyze the learning dynamics of the induced operator $Q$ under gradient flow with a squared Frobenius loss, $\mathcal{L} = \frac{1}{2} \|Q - Q^\star\|_F^2$, where $Q^\star$ is the ground-truth target operator. We assume $Q$ is parameterized via matrix factorization $Q = \mathbf{U}\mathbf{V}^\top$ (representing the active subspaces of $W$ and $V$).

\textbf{Gradient Flow Dynamics.} Parameters evolve via $\dot{\theta} = -\nabla_\theta \mathcal{L}$. The evolution of the composite operator follows the product rule $\dot{Q} = \dot{\mathbf{U}}\mathbf{V}^\top + \mathbf{U}\dot{\mathbf{V}}^\top$. Substituting the gradients yields the bilinear flow equation \citep{saxe2013exact}:
\begin{equation}
    \dot{Q} = -(Q - Q^\star)\mathbf{V}\mathbf{V}^\top - \mathbf{U}\mathbf{U}^\top(Q - Q^\star).
    \label{eq:matrix_flow}
\end{equation}

Now, let the target admit an SVD $Q^\star = \sum_{i} s_i \mathbf{u}_i \mathbf{v}_i^\top$. Under standard small-initialization regimes used in practice, the learned operator aligns with these singular subspaces: $Q(t) = \sum_i c_i(t) \mathbf{u}_i \mathbf{v}_i^\top$, where $c_i(t)$ is the learned coefficient for mode $i$. Substituting the residual $Q - Q^\star = \sum_i (c_i - s_i) \mathbf{u}_i \mathbf{v}_i^\top$ into Eq.~\ref{eq:matrix_flow} and exploiting the orthonormality of singular vectors ($\mathbf{v}_i^\top \mathbf{v}_j = \delta_{ij}$),  the first term $(Q -Q^\star)\mathbf{V}\mathbf{V}^\top $ expands as:
\begin{align}
    = \sum_i (c_i - s_i) \mathbf{u}_i (\mathbf{v}_i^\top \mathbf{V})
    \mathbf{V}^\top = \sum_i (c_i - s_i) b_i^2 \mathbf{u}_i \mathbf{v}_i^\top,
\end{align}
where $b_i = \|\mathbf{V}^\top \mathbf{v}_i\|$. By symmetry, the second term simplifies similarly with $a_i = \|\mathbf{U}^\top \mathbf{u}_i\|$.  Combining terms yields the evolution of the operator:
\begin{equation}
    \dot{Q} = -\sum_i (a_i^2(t) + b_i^2(t))(c_i(t) - s_i) \mathbf{u}_i \mathbf{v}_i^\top.
\end{equation}
Crucially, no cross-terms $\mathbf{u}_i \mathbf{v}_j^\top$ ($i \neq j$) appear. The dynamics reduce to independent scalar ODEs for each mode coefficient $\dot{c}_i(t)$. This confirms that bilinear architectures naturally decouple orthogonally separable functional components, allowing specific modes to be unlearned (setting $s_k=0$) without perturbing others.

\subsection{Implications for Extrapolation}
The derived independence of singular modes implies that bilinear models learn the \textit{generating operator} of the data. This is critical for tasks requiring recursive composition (e.g., $k$-step prediction), which corresponds to computing powers of the operator $Q^k$.

Since the effective interaction operator $Q$ is symmetric, it admits an eigendecomposition $Q = \mathbf{V}\Lambda\mathbf{V}^\top$. Consequently, repeated application simplifies to $Q^k = \mathbf{V}\Lambda^k\mathbf{V}^\top$. In piecewise-linear networks, $Q$ is approximated by local affine maps $A_x$; extrapolation $Q^k$ becomes a product of mismatched matrices $\prod A_{x_t}$, leading to multiplicative error compounding. In contrast, because gradient flow in bilinear layers preserves the singular subspaces $\mathbf{V}$ globally, the model learns a stable basis aligned with the generating operator. Extrapolation requires the scalar eigenvalues $\Lambda$ to scale, which is structurally permitted by the architecture.

\section{Experiments}
We evaluate structural disentanglement across controlled algebraic tasks, unlearning benchmarks, and long-horizon extrapolation. All architectures are matched in parameter count and optimization settings.

\subsection{Modular arithmetic}
For modular arithmetic, we work with $\mathbb{Z}_p$, where $p$ is a prime number. Here, we take $p = 97$. The models are trained to predict $(a+b) \text{ mod }p$ under modular addition and $(ab)\text{ mod }p$ under modular multiplication, given the input pair $(a, b) \in \{0, 1, \dots, p-1\}^2$.

Each model maps input pairs $(a,b)$ to class logits, from which we extract a class-specific interaction matrix $M_k(a,b)=\ell_k(a,b)$. For bilinear models, this interaction admits an explicit low-rank factorization; for pointwise and gated models, $M_k$ is obtained by evaluating logits over all input pairs. Architectural details are identical across tasks and provided in the appendix \ref{app:model_details_mod_arith}

\textbf{Fourier analysis in addition.} For modular addition, the ground-truth operator is circulant and therefore diagonalizable by the discrete Fourier transform. As a result, each output class corresponds to a fixed set of Fourier modes with uniform magnitude across frequencies. To assess whether a model recovers this structure, we compute the 2D DFT of each learned interaction matrix $M_k$ and measure spectral concentration via the Shannon entropy of the normalized power spectrum, which we term \textbf{Fourier entropy}:
\begin{equation}
    H_k = - \sum_{u,v} P_k(u,v)\,\log P_k(u,v),
\end{equation}
where 
\[
P_k(u,v) = \frac{|\widehat{M}_k(u,v)|^2}{\sum_{u',v'} |\widehat{M}_k(u',v')|^2}
\]
is the normalized power spectrum of the operator. We compare the mean entropy of the learned operators against the theoretical ground truth $(H_k)_\text{true} = \log p$, which corresponds to the uniform power distribution of the true addition operator (see Appendix~\ref{app:mod_arith_metrics}).

\textbf{Singular-value spectra in multiplication.}
In the case of multiplication, the true operator is not diagonal in the Fourier basis, but it still exhibits a structured and relatively low-rank structure. For each class $k$, we first center $M_k$ by subtracting its row and column means, in order to compute its singular values $\sigma_1 \geq \sigma_2 \geq \dots$. We visualize the normalized singular-value decay $\sigma_i / \sigma_1$, to compare how quickly the performance decays.

\subsection{Extrapolation in cyclical reasoning task}
\label{subsec:exp_extrapolation}

For the extrapolation task, we train the models to predict the value of a function $f(a)$ and evaluate their accuracy on the $i$-th iterate $f_i$, in order to analyze how well the models extrapolate. We focus on the successor function $f(a) = (a+1) \pmod p$ on a cycle of length $p$.

We define $a$ as the input integer and $\phi$ as a function identifier. To perform extrapolation, we treat the model as learning a stochastic transition matrix. We calculate the conditional probability assigned to class $k$ given input $a$: $P(k \mid a, \phi) = \mathrm{softmax}(\ell(a,\phi))_k$

We collect these values into a $p \times p$ transition operator $T$ by setting $T[k,a] = P(k \mid a, \phi)$. Each column of $T$ represents the probability distribution over output states given an input state. We analyze its sparsity by calculating the Shannon entropy of its columns. Columns that are close to one-hot (representing a deterministic permutation) possess low entropy, while diffuse columns indicate uncertainty.

Finally, we obtain predictions for the $i$-th iterate by computing higher powers of the extracted operator, $T^i$. If the model has learned the true topology of the cycle, $T$ approximates a permutation matrix, and $T^i$ correctly performs $i$ shifts. We make multi-step predictions as $\hat{f}_i(a) = \arg\max_k T^i[k,a]$ and quantify the extrapolation capability by measuring the decay of $\text{Accuracy}(i)$ as the composition depth $i$ increases. The mathematical formulations for ground truth iterates, accuracy, and entropy are detailed in Appendix~\ref{app:extrapolation_details}.

\subsection{Lie group dynamics}
\label{subsec:exp_lie_group}

\textbf{Rotation.}
3D rotations in physics are governed by quaternion differential equations. A compact mathematical object, known as the quaternion, is used to describe orientation in 3D space. It is denoted as $q = (s, \mathbf{v}) = q(w, x, y, z)$; where \( s \in \mathbb{R} \) is its scalar part and \( \mathbf{v} \in \mathbb{R}^3 \) is the vector part. An important property of a rotation quaternion is that it must always have unit norm ($\|q\| = 1$).

We train the models to predict the next state $q_{t+1}$ given the current state $q_t$ and angular velocity $\omega$. The underlying kinematics are governed by the interaction: $\dot{q} = \frac{1}{2} q \otimes \omega$;
where $\otimes$ represents the Hamilton product. We train using mean squared error on the next state without imposing a normalization constraint. We then test the \textit{emergent} long-term stability by performing iterative multi-step predictions starting from a random orientation. If the model has learned a valid representation then the norm should remain invariant at 1.

\textbf{Volume preservation.} 
The laws of incompressible fluid dynamics are governed by the Special Linear group $SL(2, \mathbb{R})$. The core concept is the Lie Algebra $\mathfrak{sl}(2, \mathbb{R})$, the set of all $2 \times 2$ real matrices with vanishing trace. This dynamical system is governed by the differential equation: $\frac{d\mathbf{x}}{dt} = \mathbf{G}\mathbf{x}$. 

The exact solution to this differential equation is $\mathbf{x}(t) = e^{\mathbf{G}t} \mathbf{x}(0)$. A fundamental property of matrix exponentials relates the trace to the determinant (Jacobi's formula): $\det(e^{\mathbf{G}t}) = e^{\text{Tr}(\mathbf{G})t} = e^{0} = 1$.

Because the determinant is 1, the transformation preserves area. We train models to approximate the first-order update rule of this system. To evaluate structural disentanglement, we check the determinant drift using the edges of a unit square defined by $v_1 = [1, 0]^\top$ and $v_2 = [0, 1]^\top$. These vectors are fed recursively into the model. At every step, the area is calculated as: $\text{Area}_t = x_{v1} y_{v2} - x_{v2} y_{v1}.$ For more details refer Appendix~\ref{app:lie_group_details}.

\section{Results}
\label{sec:results}
\begin{figure*}[t]
    \centering
    % --- ROW 1: ALPHA = 0.0 (Orthogonal) ---
    % Multiplicative Family
    \begin{subfigure}{0.16\linewidth}
        \includegraphics[width=\linewidth]{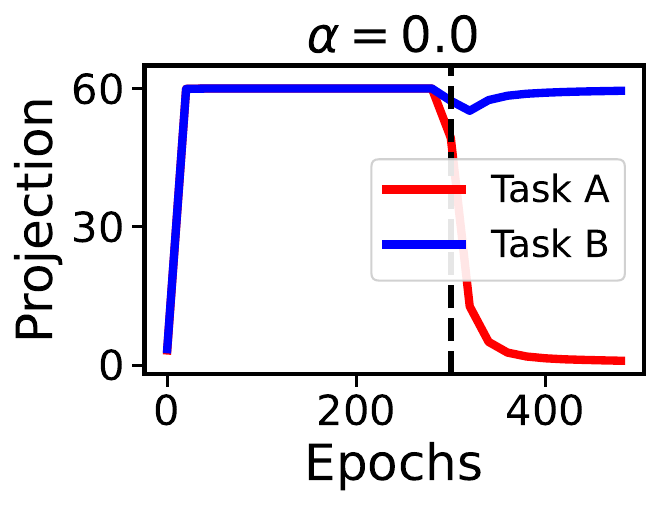}
    \end{subfigure}
    \hfill
    \begin{subfigure}{0.16\linewidth}
        \includegraphics[width=\linewidth]{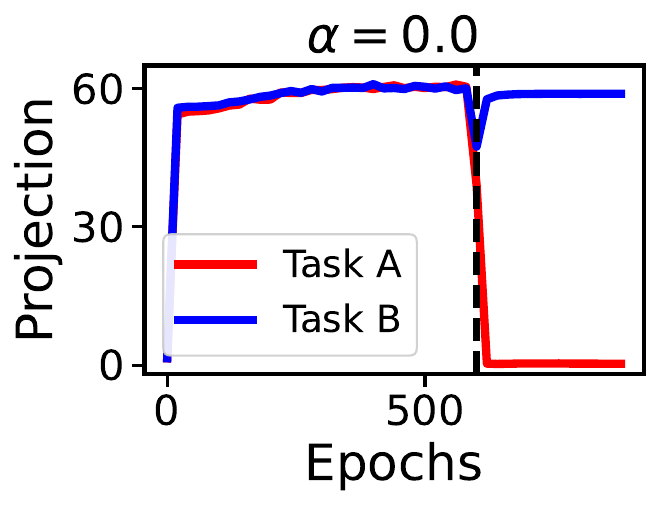}
    \end{subfigure}
    \hfill
    \begin{subfigure}{0.16\linewidth}
        \includegraphics[width=\linewidth]{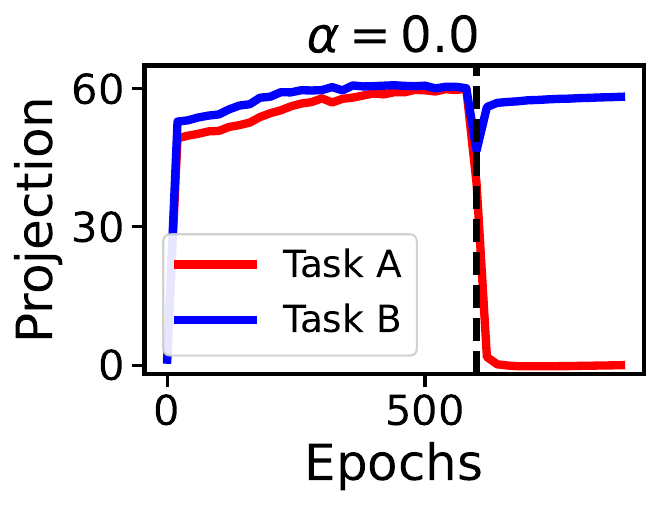}
    \end{subfigure}
    \hfill
    % Pointwise Family
    \begin{subfigure}{0.16\linewidth}
        \includegraphics[width=\linewidth]{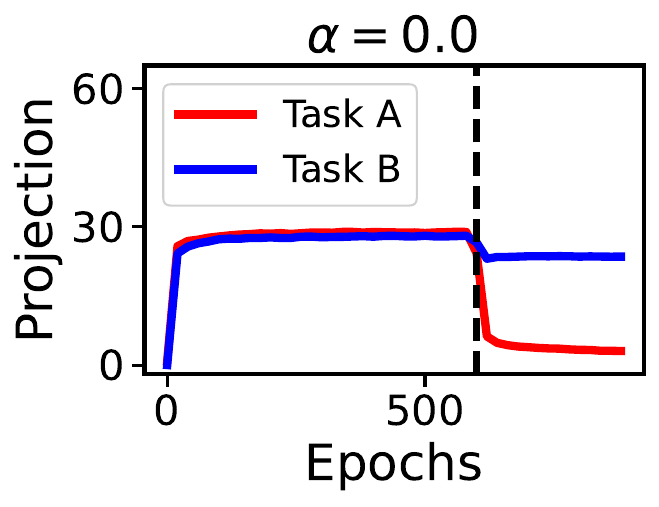}
    \end{subfigure}
    \hfill
    \begin{subfigure}{0.16\linewidth}
        \includegraphics[width=\linewidth]{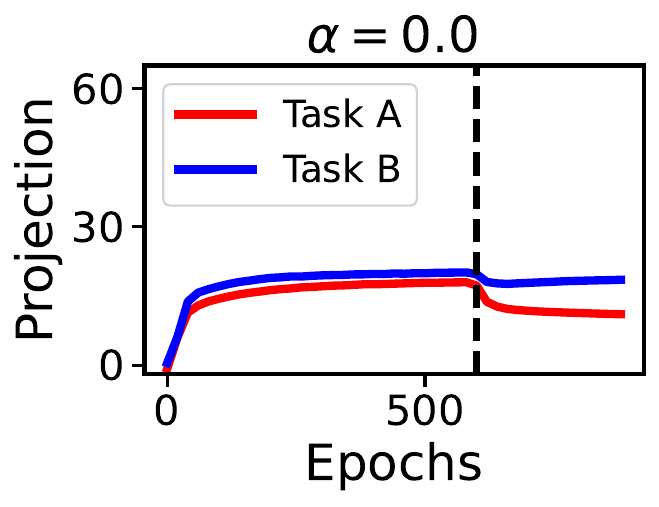}
    \end{subfigure}
    \hfill
    \begin{subfigure}{0.16\linewidth}
        \includegraphics[width=\linewidth]{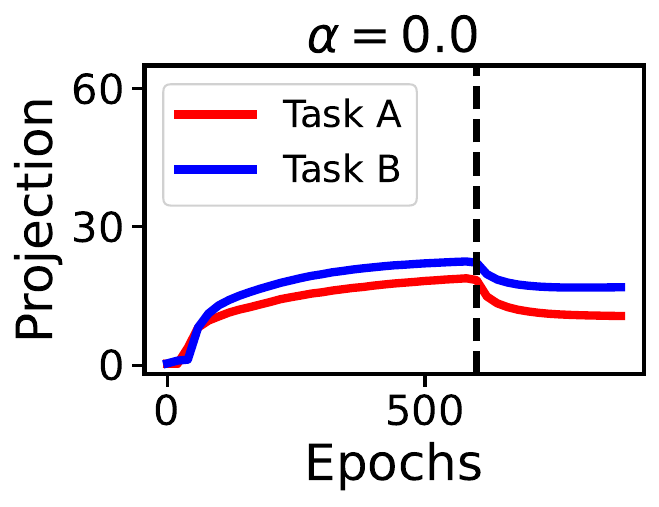}
    \end{subfigure}
    
    \vspace{0.5em} % Spacing between rows
    
    % --- ROW 2: ALPHA = 1.0 (Parallel) ---
    % Multiplicative Family
    \begin{subfigure}{0.16\linewidth}
        \includegraphics[width=\linewidth]{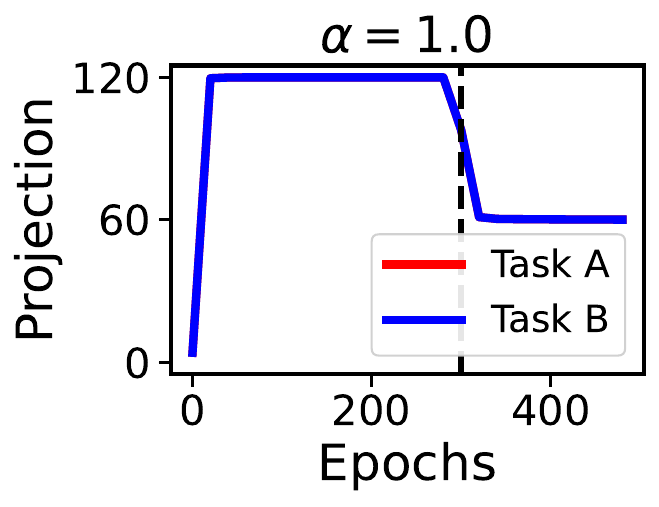}
        \caption*{\small Bilinear}
    \end{subfigure}
    \hfill
    \begin{subfigure}{0.16\linewidth}
        \includegraphics[width=\linewidth]{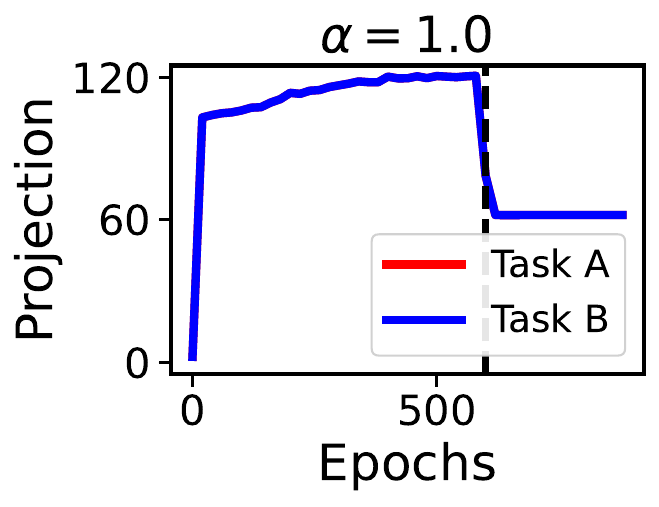}
        \caption*{\small SwiGLU}
    \end{subfigure}
    \hfill
    \begin{subfigure}{0.16\linewidth}
        \includegraphics[width=\linewidth]{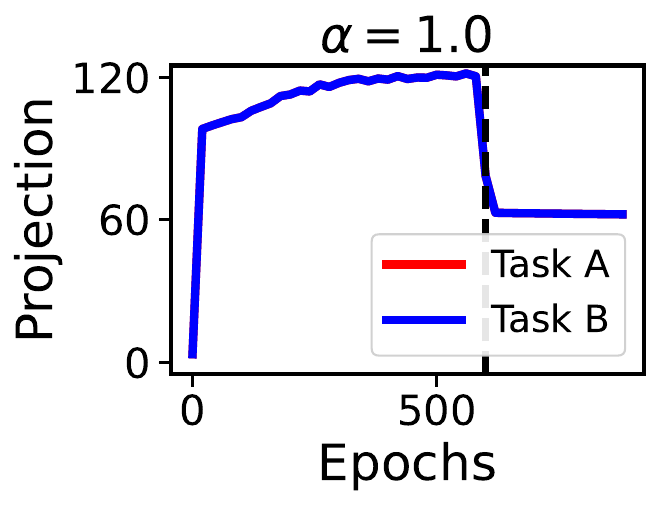}
        \caption*{\small GeGLU}
    \end{subfigure}
    \hfill
    % Pointwise Family
    \begin{subfigure}{0.16\linewidth}
        \includegraphics[width=\linewidth]{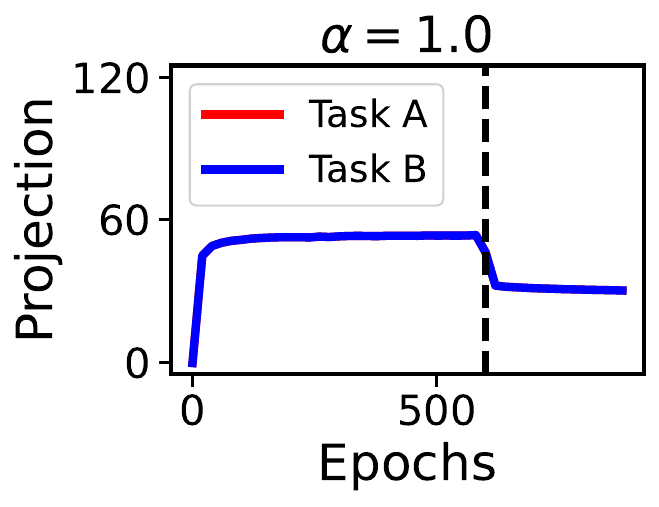}
        \caption*{\small ReLU}
    \end{subfigure}
    \hfill
    \begin{subfigure}{0.16\linewidth}
        \includegraphics[width=\linewidth]{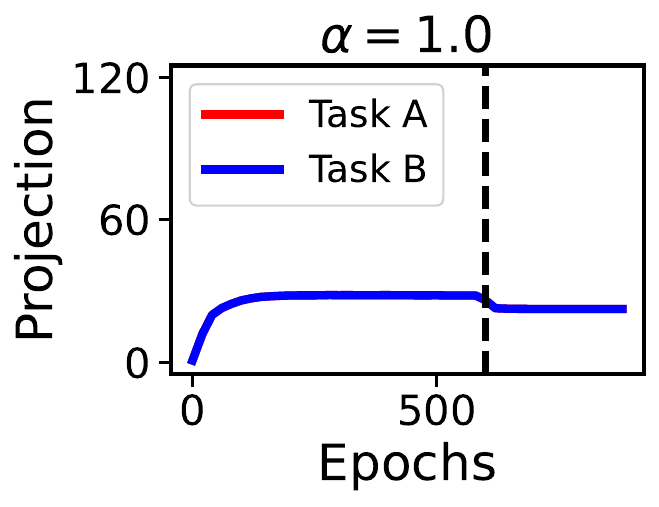}
        \caption*{\small Tanh}
    \end{subfigure}
    \hfill
    \begin{subfigure}{0.16\linewidth}
        \includegraphics[width=\linewidth]{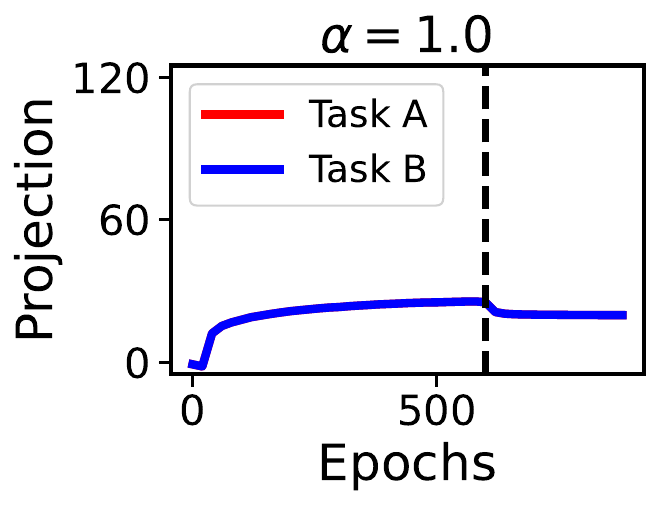}
        \caption*{\small Sigmoid}
    \end{subfigure}
    
    \vspace{-0.5em}
    \caption{\textbf{Architectural Inductive Bias: Multiplicative vs. Pointwise.} 
    Comparison of unlearning dynamics for Orthogonal ($\alpha=0$, Top) and Parallel ($\alpha=1$, Bottom) tasks.
    \textbf{Left 3 Columns (Multiplicative):} Bilinear, SwiGLU, and GeGLU architectures maintain structural separation.
    \textbf{Right 3 Columns (Pointwise):} ReLU, Tanh, and Sigmoid architectures exhibit task interference.}
    \label{fig:arch_comparison}
    \vspace{-1.3em}
\end{figure*}

We empirically investigate whether the explicit architectural inductive bias of multiplicative architectures (Bilinear MLPs, SwiGLU, GeGLU) translates into qualitatively different learning dynamics compared to standard pointwise nonlinearities (ReLU, Tanh, Sigmoid).

We organize our analysis to trace the impact of architectural inductive bias on downstream tasks. We begin by \textbf{validating} the theoretical prediction of gradient independence on a controlled synthetic benchmark. Next, we demonstrate that this independence enables \textbf{surgical unlearning} even in scenarios where tasks are entangled by shared inputs. To explain \textit{why} this separation occurs, we \textbf{diagnose} the internal representations, confirming they align with ground-truth algebraic operators. Finally, we test the stability of these operators under recursive composition: first in discrete \textbf{extrapolation} tasks, and in continuous Lie group dynamics, where we evaluate whether the model preserves physical invariants over long horizons.

\subsection{Direct Validation: Architectural Inductive Bias}
\label{subsec:gradient_validation}

Our theoretical derivation in Section~\ref{sec:theory} (Eq.~\ref{eq:matrix_flow}) shows that, for bilinear optimization, the gradient flow decomposes into independent scalar ODEs for orthogonal modes. We first validate whether this idealization holds under practical SGD conditions. We hypothesize that architectures with multiplicative interactions will preserve task geometry better.

To empirically verify this, we constructed a minimal working example designed to test geometric preservation. We define a pair of orthogonal basis vectors, $\{\mathbf{u}_1, \mathbf{u}_1^{\perp}\}$ and $\{\mathbf{v}_1, \mathbf{v}_1^{\perp}\}$, in a high-dimensional space ($\mathbb{R}^{32}$). Task A is fixed to the primary direction: $f_1(\mathbf{x}) = (\mathbf{u}_1^\top \mathbf{x})(\mathbf{v}_1^\top \mathbf{x})$. For Task B, $f_2(\mathbf{x}) = (\mathbf{u}_2^\top \mathbf{x})(\mathbf{v}_2^\top \mathbf{x})$, we construct its vectors by interpolating between the orthogonal and parallel directions using a mixing parameter $\alpha \in [0, 1]$:
\[ \mathbf{u}_2 = (1-\alpha) \cdot \mathbf{u}_1^{\perp} + (\alpha) \cdot \mathbf{u}_1 \]
This setup allows us to precisely control the cosine similarity (correlation) between the tasks, from perfectly orthogonal ($\alpha=0$) to fully parallel ($\alpha=1$). The models are first trained on the superposition $y = f_1(\mathbf{x}) + f_2(\mathbf{x})$, forcing them to represent two distinct rules. To assess disentanglement, we subsequently switch the training objective to target strictly Task 2 ($y_{new} = f_2(\mathbf{x})$), explicitly penalizing the retention of Task 1.

Figure~\ref{fig:arch_comparison} presents the unlearning dynamics. In the \textbf{Orthogonal regime ($\alpha=0$, Top Row)}, the Multiplicative models demonstrate distinct structural separation, suppressing Task A while keeping Task B strictly invariant at the target magnitude ($\sim 60$). In contrast, the pointwise models saturate at a significantly lower projection value ($\sim 30$). While the task vectors $u_1$ and $u_2$ are orthogonal in the input space, they do not map to orthogonal subspaces within a ReLU network. Because pointwise networks approximate multiplicative interactions by summing over shared neurons, they are forced to learn a distributed, polysemantic representation. 

In the \textbf{Parallel regime ($\alpha=1$, Bottom Row)}, the tasks are geometrically identical, differing only in the required output magnitude. The Multiplicative models track this additive structure closely: the projection reaches $\sim 120$ during training and drops to $\sim 60$ during unlearning. In contrast, the Pointwise models saturate at a lower magnitude ($\sim 50$). This limitation suggests that they encode signal strength and feature geometry in the same set of weights. Unlike bilinear forms where magnitude can be factored out, a ReLU network cannot reduce the output amplitude of a feature without shrinking the weights that define it.

\textbf{Quantifying Structural Distortion.}
To rigorously measure this behavior across the entire spectrum, we define \textit{Task Preservation Distortion} as the absolute deviation of the retained task's projection from the ideal target value (See Figure~\ref{fig:distortion_analysis})  (Target varies with $\alpha$: 60 at $\alpha=0$, 120 at $\alpha=1$). The Multiplicative family maintains near-zero distortion at $\alpha=0$, confirming that orthogonal tasks are treated as independent components. Pointwise models exhibit high distortion immediately at $\alpha=0$ due to the amplitude deficit.

\subsection{Surgical Unlearning in Entangled Scenarios}

To rigorously test unlearning in the presence of structural interference, we construct a \textbf{Rank-$r$ Entangled Superposition} task. The input space is partitioned into three subspaces $\mathbf{x} = [x_1; x_2; x_3] \in \mathbb{R}^{3d}$. The target is the sum of two interactions that share the central component $x_2$:
\begin{equation}
    y = \underbrace{x_1^\top A x_2}_{f_{12} \text{ (Forget)}} + \underbrace{x_2^\top B x_3}_{f_{23} \text{ (Retain)}}.
\end{equation}
This setup creates a challenging assignment problem: because $x_2$ is active in both terms, a naive model often would learn polysemantic features that respond to $x_2$ generally, rather than separating the specific interactions $(x_1, x_2)$ and $(x_2, x_3)$. We assess whether the architecture can structurally isolate the identifiable interaction $f_{12}$ for removal. % \tbc{tell why did we not report other activations for this particular one}.

\begin{figure*}[t!]
    \centering
% -----------------------------
    % Top Left: Neuron Roles
    \begin{subfigure}[t]{0.24\linewidth}
        \includegraphics[width=\linewidth]{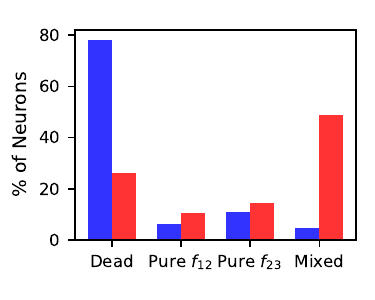}
        \label{fig:neuron_roles}
    \end{subfigure}
    \hfill
    % Top Right: Pareto
    \begin{subfigure}[t]{0.24\linewidth}
        \includegraphics[width=\linewidth]{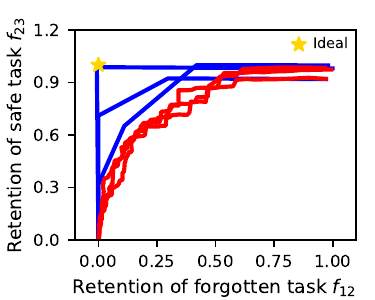}        \label{fig:pareto_unlearning}
    \end{subfigure}
    \hfill
    % Bottom Left: Gradient
    \begin{subfigure}[t]{0.24\linewidth}
        \includegraphics[width=\linewidth]{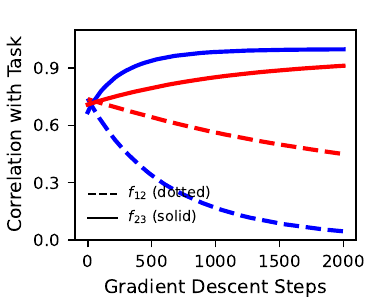}
        \label{fig:gradient_dynamics}
    \end{subfigure}
    \hfill
    % Bottom Right: Selectivity
    \begin{subfigure}[t]{0.24\linewidth}
        \includegraphics[width=\linewidth]{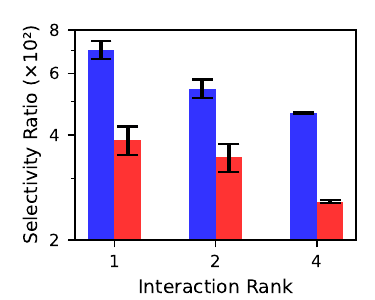}
    \end{subfigure}
    \vspace{-1.4em}
    \caption{\textbf{Surgical Unlearning Analysis.} Detailed breakdown of how \textcolor{BilinearBlue}{\textbf{Bilinear models (Blue)}} outperform \textcolor{ReluRed}{\textbf{ReLU models (Red)}} in entangled scenarios. (a) Neuron specialization prevents interference. (b) Pruning follows an ideal Pareto frontier. (c) Gradient descent moves orthogonally. (d) Selectivity remains high even as interaction rank increases.}
    \label{fig:unlearning_panel}
    \vspace{-1.5em}
\end{figure*}

\textbf{Neuron Specialization:}
We classified the functional role of each neuron by analyzing the norm of its weights corresponding to the input slots $[x_1, x_2, x_3]$. For bilinear models, this is computed as the product of the $\ell_2$ norms of the corresponding input blocks in the two multiplicative projections; for ReLU models, roles are inferred from the relative $\ell_2$ norms of the neuron’s input weights across blocks. As shown in Figure~\ref{fig:unlearning_panel}(a), the Bilinear model spontaneously enforces specialization: neurons categorize strictly as Pure $f_{12}$ (active only on $x_1, x_2$) or Pure $f_{23}$ (active only on $x_2, x_3$), with a large proportion remaining Dead (sparse). This separation arises because the bilinear product $u.x \odot v.x$ allows the network to zero out interactions involving $x_3$ when learning $f_{12}$. In contrast, the ReLU model allocates the majority of its capacity to Mixed neurons that are active across all three inputs. This confirms that ReLU struggles to decouple the shared variable $x_2$, leading to functional entanglement.

\textbf{Pareto Frontier of Pruning.}
To quantify separability, we ranked neurons by their contribution to the forgotten task $f_{12}$ and progressively pruned them. Figure~\ref{fig:unlearning_panel}(b) plots the retention of the safe task $f_{23}$ against the retention of $f_{12}$. The Bilinear model traces a trajectory near the ideal top-left corner: it is possible to remove nearly 100\% of the $f_{12}$ signal while maintaining $>90\%$ of the $f_{23}$ accuracy. The ReLU model exhibits a diagonal collapse: pruning the weights most essential for $f_{12}$ causes an immediate, proportional degradation of $f_{23}$, showing that the shared mixed neurons are essential for both tasks.

\textbf{Gradient Dynamics.}
We examined the stability of unlearning under optimization. We applied a fine-tuning objective $\mathcal{L}_{retain} = \text{MSE}(\hat{y}, f_{23})$ to the pretrained models, implicitly requiring the erasure of $f_{12}$.
The Bilinear model rapidly decouples: the correlation with $f_{12}$ (dotted line) drops to near zero within 500 steps, while the integrity of $f_{23}$ (solid line) remains preserved. In the ReLU model, the $f_{12}$ signal persists stubbornly (dotted line remains high), indicating that the gradient updates for the retained task do not orthogonalize against the forgotten task as quickly.

\textbf{Robustness and Selectivity.}
Finally, we measured the \textit{Selectivity Ratio} (defined as the ratio of damage to the forgotten task versus damage to the retained task ($\Delta f_{12} / \Delta f_{23}$) under an adversarial unlearning objective ($\mathcal{L} = \mathcal{L}_{f_{23}} - 0.5 \mathcal{L}_{f_{12}}$).
We swept the interaction rank from $r=1$ to $r=4$. As shown in Figure~\ref{fig:unlearning_panel}(d), the Bilinear architecture maintains a selectivity ratio significantly higher than the ReLU baseline across all ranks. While performance naturally degrades as the interaction complexity increases (Rank 4), the multiplicative inductive bias consistently offers a safer margin for editing.

\label{subsec:unlearning}

\begin{figure}[ht!]
    \centering
    \includegraphics[width=0.8\linewidth]{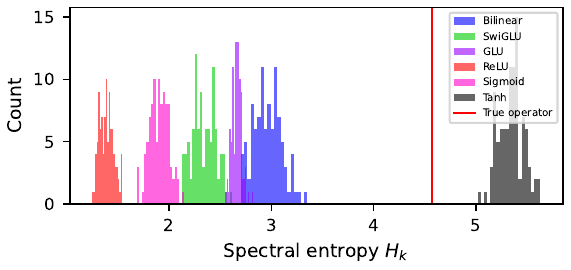}
    \caption{Spectral entropy $H_k$ across output classes for trained models. The vertical red line denotes the true addition operator. Multiplicative architectures concentrate closer to the true value, while pointwise nonlinearities exhibit either lower entropy from procedural memorization (ReLU, Sigmoid) or higher entropy from diffuse spectral representations (Tanh).}
    \label{fig:add-entropy}
    \vspace{-1.5em}
\end{figure}

% \begin{figure}[t]
%     \centering
%     \begin{minipage}{0.49\linewidth}
%         \centering
%         \includegraphics[width=\linewidth]{figures/svd_mul_k20.pdf}
%     \end{minipage}
%     \begin{minipage}{0.49\linewidth}
%         \centering
%         \includegraphics[width=\linewidth]{figures/svd_mul_k80.pdf}
%     \end{minipage}
%     \caption{Normalized singular value decay for interaction matrices in modular multiplication. The bilinear spectra (blue) decay sharply, indicating the operators are effectively lower-dimensional.}
%     \label{fig:mul-svd}
%     \vspace{-1.5em}
% \end{figure}

\begin{figure}[t]
    \centering
    \includegraphics[width=\linewidth]{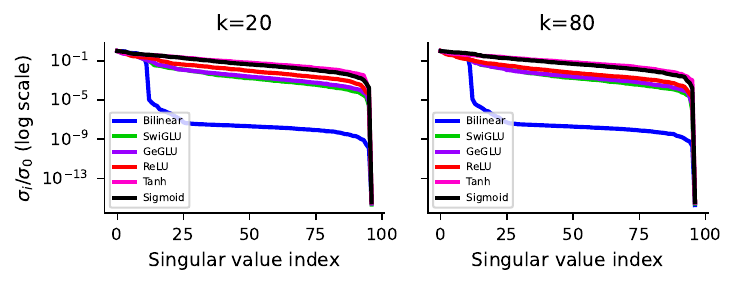}
    \caption{Normalized singular value decay for interaction matrices in modular multiplication. The bilinear spectra (blue) decay sharply, indicating the operators are effectively lower-dimensional.}
    \label{fig:mul-svd}
    \centering
    \includegraphics[width=\linewidth]{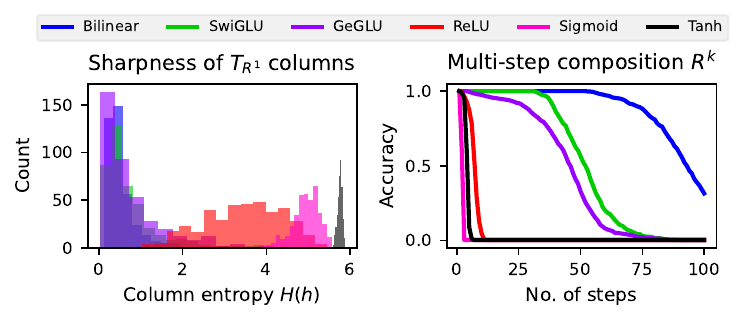}
    \vspace{-1.5em}
    \caption{Left: Distribution of column entropies for the learned transition operator $T$. Right: Multi-step composition behavior.}
    \vspace{-1.7em}
    \label{fig:cycle-entropy}
\end{figure}

\subsection{Internal Representation of Algebraic Structure}
\label{subsec:modular_arithmetic}

We examine Modular Arithmetic over $\mathbb{Z}_{97}$ to determine if the structural disentanglement observed in toy settings translates to the discovery of ground-truth algebraic operators in group-theoretic tasks. Results on $\mathbb{Z}_{113}$ are in Appendix \ref{appendix:mod-113}.

\textbf{Addition (Figure \ref{fig:add-entropy}).}
Modular addition corresponds to a circulant operator, which is diagonalized by the Discrete Fourier Transform (DFT). A model that internalizes the group operation must mimic this spectral structure.

Figure~\ref{fig:add-entropy} quantifies this via Spectral Entropy ($H_k$). The Bilinear model cluster in a structured regime ($H \approx 3.5$), being close to the theoretical ideal. Conversely, ReLU and Sigmoid collapse to extremely low entropy ($H < 2.0$). This indicates that these models memorize a few isolated frequency components to fit the training data. Interestingly, Tanh exhibits high entropy due to learned noise, reflecting a failure to compress the operator onto the correct manifold. The heatmaps are plotted in the Appendix Figure \ref{fig:add-heatmaps}.

\textbf{Multiplication and Low-Rank Factorization (Figure~\ref{fig:mul-svd}).}
Modular multiplication is inherently rank-structured. We assessed this by computing the singular value decomposition (SVD) of the centered interaction matrices.
Consistent with the additive results, the Multiplicative architectures exhibit a steep spectral decay, requiring only $\approx 25$ dimensions to capture $90\%$ of the interaction energy. This confirms the Bilinear models exploit their inductive bias to find a low-rank factorization of the multiplication table. 

% \begin{figure}[h]
    
% \end{figure}

\subsection{Discrete Extrapolation: Cyclic Reasoning}
\label{subsec:extrapolation}

A model that truly learns the generating operator of the data should remain stable under recursive composition (iterating the function $f(x)$ as $f(f(x)) \dots$). We test this hypothesis on a simple long-horizon extrapolation task.

\textbf{Discrete Cyclic Reasoning.}
We modeled cyclic dynamics on a graph of $N=400$ nodes. The task requires learning the successor relation $i \to (i+1) \pmod N$. We extracted the learned transition operator $T$ and evaluated its stability under recursive application ($T^k$).
Since the ground truth transformation is a permutation, a valid model should maintain accuracy for large $k$. The Multiplicative models (Bilinear, SwiGLU, GeGLU) learned highly sparse transition matrices that closely approximated a permutation (low mean column entropy, see Figure~\ref{fig:cycle-entropy}). Consequently, the Bilinear model maintained predictive accuracy for $k \approx 80$ steps of composition, followed closely by GeGLU and SwiGLU. In contrast, the Pointwise family (ReLU, Tanh, Sigmoid) collapsed almost immediately ($k < 10$). Their internal operators were diffuse, noisy approximations (high entropy) which compounded exponentially during iteration.

\subsection{Continuous Extrapolation: Lie Group Dynamics}
\label{subsec:lie_groups}

We extend this analysis to continuous symmetries governed by Lie groups. In physical systems, a valid model must preserve invariants to avoid integration drift.

\begin{figure}[t!] 
    \centering

    % single combined image (no duplication)
    \includegraphics[width=\linewidth]{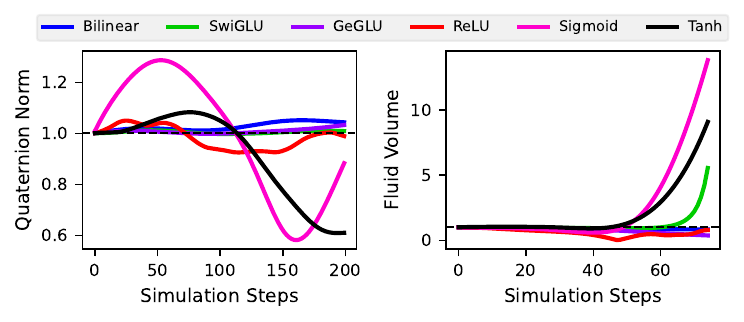}

    \begin{subfigure}{0.48\linewidth}
    \end{subfigure}
    \hfill
    \begin{subfigure}{0.48\linewidth}
    \end{subfigure}
    \vspace{-1.7em}
    \caption{Long-horizon extrapolation on Lie Group dynamics. \textbf{Left}: Rigid Body Stability ($S^3$): The multiplicative models preserve the unit norm constraint better. \textbf{Right}: Fluid Volume Conservation ($SL(2)$): The multiplicative models preserve the determinant near 1.0 better.}
    \label{fig:physics_stability}
    \vspace{-2em}
\end{figure}

\textbf{Rigid Body Dynamics ($S^3$):} We simulated 3D rotations governed by unit quaternions. As shown in Figure~\ref{fig:physics_stability} (a), the multiplicative models adhere tightly to the unit sphere with negligible variance over 200 steps. In contrast, Pointwise models exhibit chaotic instability; Tanh oscillates wildly while Sigmoid loses the manifold entirely.

\textbf{Volume Preserving Flows ($SL(2)$):} We simulated incompressible fluid deformation where the determinant must remain $1.0$. Figure~\ref{fig:physics_stability} (b) reveals a stark difference in error dynamics. The multiplicative models exhibit improved conservation (flat lines near 1.0). Conversely, volume calculated using Pointwise models expands with increasing steps. This confirms that  piecewise-linear approximations fail to respect the volume-preserving constraint.

\section{Discussion and Limitations}
This work argues that the feasibility of selective unlearning and long-horizon extrapolation is fundamentally constrained by how a model structures its internal representations upon learning. Our results suggest that, for tasks governed by compositional or algebraic structure, architectural inductive bias plays a key role in determining whether learned functions decompose into independently separable components.

A central implication of this work is that unlearning is a \textit{representational problem rather than an algorithmic one}. Our results show that when architectural bias enforces structural disentanglement during learning, unlearning becomes stable, independent of the specific unlearning procedure applied. This view also clarifies the connection between unlearning and extrapolation observed across our experiments. In both cases, the model must preserve a learned operator while selectively suppressing or scaling specific components: unlearning removes a subset of functional modes, while extrapolation repeatedly composes the same operator without distortion. Architectures that entangle functional components cause interference in representational space. But when the learned operator decomposes into orthogonal subspaces, both selective removal and repeated composition become structurally well-posed.

Importantly, we don't propose bilinear MLPs as universally superior architectures. Rather, they serve as a controlled setting in which multiplicative interactions are explicitly reflected in the learned operator. The observed advantages arise specifically in regimes where the target function admits an underlying compositional structure. In these settings, bilinear parameterizations keep modes independent during optimization, while pointwise nonlinear architectures approximate similar functions using entangled representations.

More broadly, these findings suggest that architectural inductive bias should be treated as an important consideration when designing models intended to support editing, unlearning, or systematic generalization. Post-hoc algorithmic fixes only help when the learned representation already has the appropriate structure. Understanding and enforcing this structure at training time may therefore be a prerequisite for reliable model editing or for continually evolving systems.

\textbf{Limitations:} Our analysis focus on settings where the target function admits an identifiable compositional or algebraic structure, such as modular arithmetic, cyclic dynamics, or Lie group dynamics. As these are intentionally chosen to make structural properties analyzable, the results do not directly imply that similar disentanglement will emerge in unstructured perceptual tasks or open-ended domains.

The theoretical analysis relies on idealized assumptions, including gradient flow dynamics, squared loss, and small initialization, which allow the non-mixing property of bilinear parameterizations to be derived analytically. Although our empirical results demonstrate that the qualitative predictions persist under practical training regimes, we do not provide formal guarantees beyond this idealized regime.

Finally, bilinear MLPs are used as a minimal and analytically tractable probe of multiplicative inductive bias, rather than as a prescriptive architecture for all applications. Other architectures with partial or implicit multiplicative structure may exhibit related behavior, but characterizing the bare-minimum architectural bias sufficient to induce structural disentanglement remains an open problem.

\section{Conclusion}
This work examined how architectural inductive bias shapes the internal structure of learned representations and, thus, determines how selective unlearning and extrapolation are feasible. We showed that when a model’s learned operator decomposes into orthogonal components, functional modes can be modified or composed without interference. Across a range of settings, we observe a consistent pattern: architectures with explicit multiplicative interactions recover operators aligned with the ground-truth structure, while pointwise nonlinear networks tend to rely on polysemantic representations despite achieving similar training accuracy. This difference appears in model behavior: bilinear models support surgical unlearning with minimal collateral damage and remain stable under repeated composition, whereas pointwise baselines struggle under both criteria. Taken together, these results suggest that model editability and generalization are fundamentally constrained by representational structure learned during training. Architectural inductive bias, rather than post-hoc algorithms alone, plays a central role in determining whether such structure emerges.

\appendix

\section*{Impact Statement}
This paper presents work whose goal is to advance the field of Machine
Learning. There are many potential societal consequences of our work, none
which we feel must be specifically highlighted here.


\bibliography{main}
\bibliographystyle{icml2026}

%%%%%%%%%%%%%%%%%%%%%%%%%%%%%%%%%%%%%%%%%%%%%%%%%%%%%%%%%%%%%%%%%%%%%%%%%%%%%%%
%%%%%%%%%%%%%%%%%%%%%%%%%%%%%%%%%%%%%%%%%%%%%%%%%%%%%%%%%%%%%%%%%%%%%%%%%%%%%%%
% APPENDIX
%%%%%%%%%%%%%%%%%%%%%%%%%%%%%%%%%%%%%%%%%%%%%%%%%%%%%%%%%%%%%%%%%%%%%%%%%%%%%%%
%%%%%%%%%%%%%%%%%%%%%%%%%%%%%%%%%%%%%%%%%%%%%%%%%%%%%%%%%%%%%%%%%%%%%%%%%%%%%%%
\clearpage
\appendix
\section*{Experimental Details}
\section{Modular Arithmetic}

\subsection{Model Details}
\label{app:model_details_mod_arith}
Input integers $a, b$ are passed through an embedding matrix to get $e_a, e_b$.

\paragraph{Bilinear.}
The hidden state $h$ is computed as:
\begin{equation}
    h = (e_a W_1) \odot (e_bW_2)
\end{equation}
where $W_1, W_2 \in \mathbb{R}^{d \times m}$ and $h \in \mathbb{R}^m$.

\paragraph{GLU variants.}
The input embeddings $e_a$ and $e_b$ are concatenated to form $x = [e_a; e_b]$.
\begin{align}
    u &= x W_1, \\
    v &= x W_2, \\
    h &= u \odot \sigma (v). 
\end{align}
where $W_1, W_2 \in \mathbb{R}^{2d \times m}$ and $h \in \mathbb{R}^m$. 

In case of both bilinear and its GLU variants, the output logits $\ell$ and probabilities $\hat{y}$ are:
\begin{align}
    \ell &= h W_{\text{out}}, \\
    \hat{y} &= \mathrm{softmax}(\ell).
\end{align}

\paragraph{Standard MLP.} 
The input embeddings are concatenated to form $x = [e_a; e_b]$. The hidden state and output are computed as:
\begin{align}
    h &= \sigma(W x + b), \\
    \ell &= h W_{\text{out}}, \\
    \hat{y} &= \mathrm{softmax}(\ell),
\end{align}
with $W \in \mathbb{R}^{m \times 2d}$.

\begin{table}[h]
    \centering
    \caption{Hyperparameters for Modular Arithmetic Experiments}
    \label{tab:hyperparams_mod_arith}
    \begin{tabular}{l c}
        \toprule
        \textbf{Hyperparameter} & \textbf{Value} \\
        \midrule
        Modulus ($p$) & 97 \\
        Embedding Dimension ($d$) & 32 \\
        Hidden Dimension ($m$) & 64 \\
        Dataset Size & 9409 \\
        Train/Validation Split & 90\% / 10\% \\
        Batch Size & 256 \\
        \midrule
        \multicolumn{2}{c}{\textit{Optimization}} \\
        \midrule
        Optimizer & AdamW \\
        Learning Rate & $1 \times 10^{-3}$ \\
        Weight Decay & 0.1 \\
        Max Epochs & 2000 \\
        Early Stopping Threshold & 99.9\% Val Acc \\
        \midrule
        \multicolumn{2}{c}{\textit{Initialization}} \\
        \midrule
        Embeddings & Default \\
        Linear Layers & Default \\
        \bottomrule
    \end{tabular}
\end{table}

\subsection{Analysis Metrics}
\label{app:mod_arith_metrics}

\paragraph{Fourier Analysis Formulation.}
We perform a two-dimensional discrete Fourier transform (DFT) on each $M_k$ using:
\begin{equation}
    \widehat{M}_k = F M_k F^\ast,
\end{equation}
where $F$ is the $p \times p$ DFT matrix and $\ast$ denotes the conjugate transpose operation. We then form a normalized power spectrum to obtain a probability distribution over frequency pairs $(u, v)$:
\begin{equation}
    P_k(u,v) = \frac{\left|\widehat{M}_k[u,v]\right|^2}{\sum_{u',v'} \left|\widehat{M}_k[u',v']\right|^2}.
\end{equation}
We define Fourier entropy as the Shannon entropy of this normalized power distribution:
\begin{equation}
    H_k = - \sum_{u,v} P_k(u,v)\,\log P_k(u,v).
\end{equation}
For the true addition operator, $P_k$ is uniform over all $p$ bins, so $(H_k)_\text{true} = \log p$.

\section{Discrete Extrapolation: Cyclic Reasoning}
\label{app:extrapolation_details}

\subsection{Task Setup and Model Implementation}
The task involves an input integer $a$ and a function identifier $\phi$ (where $\phi \in \{0\}$ for the successor function). We use an embedding matrix $E$ to obtain $e_a = E[a]$ and $e_\phi = E[\phi]$.

\paragraph{Bilinear.}
The model computes the logits $\ell$ via separate projections, preserving the multiplicative path:
\begin{equation}
    \ell = \bigl( (e_a W_a) \odot (e_\phi W_\phi) \bigr)\, W_{\text{out}}.
\end{equation}

\paragraph{GLU Variants.}
The embeddings are concatenated and then passed through the GLU layers
\begin{align}
    x &= [e_a; e_b]\\
    \ell &= \bigl( (x W_1) \odot \sigma(y W_2) \bigr)\, W_{\text{out}}
\end{align}

\paragraph{Standard MLP.}
The model concatenates embeddings and processes them through a standard non-linear layer:
\begin{equation}
    \ell = \sigma\!\left(W [e_a; e_\phi] + b\right) W_{\text{out}}.
\end{equation}

\subsection{Evaluation Metrics}

\paragraph{Ground Truth Iterates.}
Using the properties of modular arithmetic, the ground truth value for the $i$-th iterate is defined as:
\begin{equation}
    f_i(a) = (a + i) \pmod p.
\end{equation}

\paragraph{Transition Operator Entropy.}
To quantify the sharpness of the learned transition operator $T$, we compute the Shannon entropy for each column $a$ (input state):
\begin{equation}
    H(a) = - \sum_{k=0}^{p-1} T[k,a] \log T[k,a].
\end{equation}
A value of $0$ indicates a deterministic (one-hot) transition, while $\log p$ indicates a uniform distribution (random guess).

\paragraph{Accuracy Calculation.}
After computing the $i$-th power of the transition operator $T^i$ and extracting the prediction $\hat{f}_i(a) = \arg\max_k T^i[k,a]$, we compute the accuracy over all possible inputs $a$:
\begin{equation}
    \text{Accuracy}(i) = \frac{1}{p} \sum_{a=0}^{p-1} \mathbb{I} \{\hat{f}_i(a) = f_i(a)\},
\end{equation}
where $\mathbb{I}$ is the indicator function.

\section{Continuous Extrapolation: Lie Groups}
\label{app:lie_group_details}

\subsection{Rotation Dynamics ($S^3$)}

\paragraph{Data Generation (Integration Rule).}
While the continuous dynamics are governed by $\dot{q} = \frac{1}{2} q \otimes \omega$, the discrete training data is generated using a first-order Euler integration step. Given a current state $q_t$ and angular velocity $\omega$, the ground truth target $q_{t+1}$ is calculated as:
\begin{equation}
    q_{t+1} = q_t + \dot{q}_t \Delta t = q_t + \left( \frac{1}{2} q_t \otimes \omega \right) \Delta t.
\end{equation}
The models are trained to predict this $q_{t+1}$ minimizing the Mean Squared Error (MSE), without any explicit regularization to enforce the unit norm constraint ($\|q\|=1$).

\paragraph{Model Architectures.}
The models take the quaternion $q \in \mathbb{R}^4$ and angular velocity $\omega \in \mathbb{R}^3$ as inputs.
\begin{itemize}
    \item \textbf{Bilinear MLP:} We apply separate linear projections $W_q \in \mathbb{R}^{4 \times h}$ and $W_\omega \in \mathbb{R}^{3 \times h}$ (where $h$ is hidden dimension). The update is computed via a bilinear interaction:
    \begin{equation}
        \Delta q = \bigl( (q W_q) \odot ( \omega W_\omega) \bigr) W_{\text{out}}.
    \end{equation}
    
    \item \textbf{GLU MLP:} We concatenate the inputs into a vector $x = [q; \omega] \in \mathbb{R}^7$ and pass it through a GLU MLP:
    \begin{equation}
        \ell = \bigl( (x W_1) \odot \sigma(y W_2) \bigr)\, W_{\text{out}}
    \end{equation}
    
    \item \textbf{Standard MLP:} Concatenate and pass it through feed-forward layers:
    \begin{equation}
        \Delta q = W_2 \cdot \sigma(W_1 x + b_1) + b_2.
    \end{equation}
\end{itemize}
In all cases, the final prediction is $\hat{q}_{t+1} = q_t + \Delta q$.

\subsection{Volume Preservation ($SL(2)$)}

\paragraph{Data Generation (Taylor Approximation).}
The system is defined by $\frac{d\mathbf{x}}{dt} = \mathbf{G}\mathbf{x}$, where $\mathbf{G} \in \mathfrak{sl}(2, \mathbb{R})$ is a traceless $2 \times 2$ matrix. While the exact solution is the matrix exponential $\mathbf{x}(t) = e^{\mathbf{G}t} \mathbf{x}(0)$, we train the models to learn the infinitesimal generator via its first-order Taylor approximation. The target for the next time step is:
\begin{equation}
    x_{t+1}^{\text{target}} = x_t + \mathbf{G} x_t \Delta t.
\end{equation}
This setup tests whether the model learns to approximate the operator $I + \mathbf{G}\Delta t$ in a way that respects the determinant constraint ($\det \approx 1$) over long horizons.

\paragraph{Model Architectures.}
The input consists of the state vector $x \in \mathbb{R}^2$ and a parameter vector $p$ that defines the generator $\mathbf{G}$.
\begin{itemize}
    \item \textbf{Bilinear}
    \begin{align}
        u &= W_x x, \quad v = W_p p \\
        \Delta x &= W_{\text{out}} (u \odot v)
    \end{align}

    \item \textbf{GLU}
    \begin{align}
        \Delta x &= W_{\text{out}} ((W_1 [x; p]\odot \sigma (W_2[x; p]))
    \end{align}
    
    \item \textbf{Standard MLP}
    \begin{align}
        h &= \sigma(W_1 [x; p] + b_1) \\
        \Delta x &= W_2 h + b_2
    \end{align}
\end{itemize}
The prediction is $\hat{x}_{t+1} = x_t + \Delta x$.

\paragraph{Evaluation Metric Details.}
To rigorously measure volume preservation, we track the deformation of a unit square. We define two basis vectors:
\begin{equation}
    v_1^{(0)} = \begin{pmatrix} 1 \\ 0 \end{pmatrix}, \quad v_2^{(0)} = \begin{pmatrix} 0 \\ 1 \end{pmatrix}.
\end{equation}
The initial area is $\det([v_1^{(0)}, v_2^{(0)}]) = 1.0$. During inference, we feed these vectors recursively into the trained model: $v^{(t+1)} = \mathcal{M}(v^{(t)})$. At step $t$, the preserved area is calculated as the determinant of the matrix formed by the two vectors:
\begin{equation}
    \text{Area}_t = x_{v1}^{(t)} y_{v2}^{(t)} - x_{v2}^{(t)} y_{v1}^{(t)}.
\end{equation}
Divergence from $1.0$ indicates a violation of the conservation law.

\begin{table*}[h]
    \centering
    \caption{Hyperparameters for Lie Algebra Experiments (SO(3) and SL(2))}
    \label{tab:lie_hyperparams}
    % Increases row height slightly for better readability
    \renewcommand{\arraystretch}{1.2}
    \begin{tabular}{l c c}
        \toprule
        \textbf{Hyperparameter} & \textbf{Rotation Dynamics (SO(3))} & \textbf{Volume Preservation (SL(2))} \\
        \midrule
        \multicolumn{3}{c}{\textit{Model Architecture}} \\
        \midrule
        Hidden Dimension ($h$) & 128 & 64 \\
        Input Dimension & $4+3=7$ ($q, \omega$) & $2+4=6$ ($x, G_{\text{flat}}$) \\
        \midrule
        \multicolumn{3}{c}{\textit{Optimization}} \\
        \midrule
        Optimizer & Adam & Adam \\
        Learning Rate & $1 \times 10^{-3}$ & $1 \times 10^{-3}$ \\
        Batch Size & 128 & 128 \\
        Training Iterations & 5000 & 5000 \\
        Loss Function & MSE & MSE \\
        \midrule
        \multicolumn{3}{c}{\textit{Data Generation}} \\
        \midrule
        Time Step ($\Delta t$) & 0.1 & 0.1 \\
        \bottomrule
    \end{tabular}
\end{table*}

% \paragraph{Effective Rank Formulation.}
% To calculate the effective rank at an energy level $\alpha \in (0,1)$, we find the smallest integer $r$ such that the following holds for the singular values $\sigma_i$ of the centered matrix $M_k$:
% \begin{equation}
%     \sum_{i=1}^r \sigma_i^2 \;\ge\; \alpha \sum_{i=1}^{p} \sigma_i^2.
% \end{equation}

\section{Unlearning Experiments}
\subsection{Validation for Architecture Inductive Bias}
For the validation of our theoretical insights, we perform an unlearning validation study as discussed. Firstly, we sample two random matrices $A, B \in \mathbb{R}^{D \times D}$, and apply QR decomposition
\begin{equation}
A = Q_U R_U,\quad B = Q_V R_V
\end{equation}
in  order to obtain orthonormal vectors 
\begin{equation}
u_1 = Q_U[:,0], \quad u_{\perp} = Q_U[:,1]
\end{equation}
\begin{equation}
v_1 = Q_V[:,0], \quad v_{\perp} = Q_V[:,1]
\end{equation}
i.e., 
\begin{equation}
u_1^\top u_{\perp} = 0, \quad \|u_1\|=\|u_{\perp}\|=1
\end{equation}
\begin{equation}
v_1^\top v_{\perp} = 0, \quad \|v_1\|=\|v_{\perp}\|=1
\end{equation}
we sample $\alpha \in (0,1)$, and calculate 
\begin{align}
u_2 &= \alpha\, u_1 + (1-\alpha)\, u_{\perp}, \\
v_2 &= \alpha\, v_1 + (1-\alpha)\, v_{\perp}.
\end{align}
followed by normalization 
\begin{align}
    u_2 = \frac{u_2}{\|u_2\|_2},  \quad v_2 = \frac{v_2}{\|v_2\|_2}
\end{align}
Then we obtain a scaled rank-1 matrix constructed from the outer product of two vectors $u$ and $v$ for each of the tasks using 
\begin{align}
 M_A = \lambda \cdot (u_1 v_1^T), \quad
 M_B = \lambda \cdot (u_2 v_2^T)
\end{align}
where $\lambda$ is a scalar.
We define each task dataset as $y = e_a^T M e_b$ for all integer inputs $a, b \in \{0,1, \dots, N-1\}$, where $e_a$ and $e_b$ are obtained from the rows of embedding matrix $E \in \mathbb{R}^{N \times d}$. 
We train each model on two the two tasks obtained using $M_1$ and $M_2$. 
Note that for $\alpha=0$ the tasks are completely orthogonal and $\alpha=1$ implies perfect overlap.
For the bilinear model, $e_a$ and $e_b$ are passed separately and it computes their bilinear interaction, whereas for other models we concatenate $e_a$ and $e_b$ and then pass it further into the model. 
\begin{align}
y_{\text{Bilinear}} &= w_{\text{out}}^\top \left( (W_1 e_a) \odot (W_2 e_b) \right) \\
y_{\text{GLU}} &= w_{\text{out}}^\top \left( \sigma(W_{\text{gate}} [e_a; e_b]) \odot (W_{\text{val}} [e_a; e_b]) \right) \\
y_{\text{MLP}} &= w_{\text{out}}^\top \sigma \left( W_{\text{in}} [e_a; e_b] + b_{\text{in}} \right)
\end{align}
where $\sigma$ denotes an activation function.
The models are trained on the mean-squared error (MSE) loss using Adam optimizer.
In the unlearning phase, we further train using only the task B using stochastic gradient descent. 
To quantify the strength of the learned representations, we extract the effective interaction matrix $M_{\text{learned}}$ from the model and find its similarity to the ground truth representations for each task using
\begin{equation}
\text{Score}_k = u_k^\top M_{\text{learned}} v_k, \quad k \in \{A, B\}
\end{equation}
Ideally, during the unlearning phase, $\text{Score}_A$ should go to $0$ (unlearning) while $\text{Score}_B$ to $\lambda$ (retention).
This interaction matrix is extracted via weight analysis for the bilinear model
\begin{equation}
    M_{\text{Bilinear}} = W_1^\top \operatorname{diag}(w_{\text{out}}) W_2.
\end{equation}
As we cannot extract the interaction for other models directly due to non-linearities, we calculate the score for them directly by probing
\begin{equation}
\text{Score}_k = w_{\text{out}}^\top \sigma \left( W_{\text{in}} [u_k; v_k] + b_{\text{in}} \right), \quad k \in \{A, B\}
\end{equation}

\begin{table*}[h]
    \centering
    \caption{Hyperparameters for Orthogonal Unlearning Experiments}
    \label{tab:hyperparams_ortho}
    \begin{tabular}{l c c}
        \toprule
        \textbf{Hyperparameter} & \textbf{Bilinear} & \textbf{Others (ReLU/SwiGLU/etc.)} \\
        \midrule
        Embedding Dimension ($d$) & 32 & 32 \\
        Hidden Dimension ($m$) & 128 & 128 \\
        Number of Tokens & 500 & 500 \\
        Dataset Size & 8000 & 8000 \\
        Batch Size & 256 & 256 \\
        Target Scale Factor ($\lambda$) & 60.0 & 60.0 \\
        \midrule
        \multicolumn{3}{c}{\textit{Optimization}} \\
        \midrule
        Phase 1 Optimizer & Adam (lr=0.005) & Adam (lr=0.005) \\
        Phase 2 Optimizer & SGD (lr=0.01) & SGD (lr=0.01) \\
        Phase 1 Epochs & 300 & 600 \\
        Phase 2 Epochs & 200 & 300 \\
        \midrule
        \multicolumn{3}{c}{\textit{Initialization}} \\
        \midrule
        Embeddings & Normal + Normalize & Normal + Normalize \\
        Linear Layers & Kaiming Normal & Kaiming Normal \\
        Output Layer & Normal ($\sigma=0.1$) & Normal ($\sigma=0.1$) \\
        \bottomrule
    \end{tabular}
\end{table*}

\subsection{Surgical Unlearning in Entangled Scenarios}
For the surgical unlearning experiment, the input vector $\mathbf{x}$ is split into three distinct segments:$$\mathbf{x} = [\mathbf{x}_1, \mathbf{x}_2, \mathbf{x}_3] \in \mathbb{R}^{d_1 \times d_2 \times d_3}$$
where sampled from a standard normal distribution: $\mathbf{x} \sim \mathcal{N}(0, 1)$

We generate low-rank matrices 
\begin{equation}
A_{12} = \mathbf{u}_{12} \mathbf{v}_{12}^\top;   A_{23} = \mathbf{u}_{23} \mathbf{v}_{23}^\top
\end{equation}
where $\mathbf{u}_{12} \in \mathbb{R}^{d_1 \times 1}$, $\mathbf{v}_{12} \in \mathbb{R}^{d_2 \times 1}$, $\mathbf{u}_{23} \in \mathbb{R}^{d_2 \times 1}$ and $\mathbf{v}_{23} \in \mathbb{R}^{d_3 \times 1}$.
Each of these matrices correspond to two separate interactions. 
The target values are generated by summing up these interactions
\begin{equation}
    f_{12} = \mathbf{x}_1^\top A_{12} \mathbf{x}_2, \quad
f_{23} = \mathbf{x}_2^\top A_{23} \mathbf{x}_3
\end{equation}
The final label $\mathbf{y}$ is simply the sum of these two independent interactions: $$\mathbf{y} = y^{(12)} + y^{(23)}$$
We train both the models on this dataset on MSE Loss using Adam optimizer with L1 regularization.

\textbf{Neuron roles.} Neuron Roles. We classify model neurons as Dead, Pure Interaction 1 ($f_{12}$), Pure Interaction 2 ($f_{23}$), or Mixed.
For bilinear, we partition weights according to input splits
\begin{equation}
    \mathbf{u}_h = [\mathbf{u}_{h1}, \mathbf{u}_{h2}, \mathbf{u}_{h3}], \quad\mathbf{v}_h = [\mathbf{v}_{h1}, \mathbf{v}_{h2}, \mathbf{v}_{h3}].
\end{equation}
Then the interaction scores are calculated as 
\begin{align}
    S_{12} = \|\mathbf{u}_{h1}\| \|\mathbf{v}_{h2}\| + \|\mathbf{u}_{h2}\| \|\mathbf{v}_{h1}\|\\
    S_{23} = \|\mathbf{u}_{h2}\| \|\mathbf{v}_{h3}\| + \|\mathbf{u}_{h3}\| \|\mathbf{v}_{h2}\|
\end{align}
$S_{12}$ measures the interactions between $\mathbf{x_1}$ and $\mathbf{x_2}$, showing the neurons for $f_{12}$. Similarly $S_{23}$ does it for $f_{23}$. Based on a threshold on $S_{ij}$ we classify the neurons as belonging to either of the tasks, being dead or mixed. 
For the ReLU model, we directly classify them based on neuron's connection magnitudes to the three input chunks.
For example, a neuron is said to be ``Pure $f_{12}$" if it is connected to both $x_1$ and $x_2$ (has a higher weights magnitude), but not $x_3$.

\textbf{Pareto frontier.} We iteratively prune neurons in order of their importance and observe the drop in performances.  
To evaluate this importance, for bilinear, we use a cross-norm logic $S_{12}$ as mentioned above.
For ReLU, it calculates $w_1 \cdot w_2$, where $w_1$ is the norm of the weights connected to the first set of inputs, and $w_2$ is the norm for the second. This identifies neurons that ``bridge" two input features.
We visualize this experiment on 3 random seeds to verify the robustness of our results.

\textbf{Gradient-based Unlearning.} After pre-training on the dataset as defined above, we add extra steps of training, but with the target being only the second interaction $f_{23}$. We measure the correlation of the predicted value with both $f_{12}$ and $f_{23}$ using a batch of model predictions and simply calculating the pearson correlation coefficient with the batch of corresponding true values of $f_{12}$ and $f_{23}$. 

\textbf{Selectivity ratio.} We perform a simultaneous gradient descent on $f_{23}$ and gradient ascent on $f_{12}$ using the loss function 
\begin{equation}
    \mathcal{L} = \mathcal{L}_{f_{23}} - 0.5 \mathcal{L}_{f_{12}}
\end{equation}
We calculate the selectivity ratio over different ranks of the interaction matrix $A_{ij}$, which is defined as ratio of drop in the values of correlation coefficient of $f_{23}$ and $f_{12}$ with the predicted value. The correlation coefficient logic is same as that explained in the previous sections.

\begin{table*}[h]
    \centering
    \caption{Hyperparameters for Robustness and Selectivity Experiments}
    \label{tab:hyperparams}
    \begin{tabular}{l c c}
        \toprule
        \textbf{Hyperparameter} & \textbf{Bilinear} & \textbf{ReLU} \\
        \midrule
        \multicolumn{3}{c}{\textit{Architecture \& Data}} \\
        \midrule
        Subspace Dimension ($d_1, d_2, d_3$) & 16 & 16 \\
        Hidden Dimension ($m$) & 64 & 64 \\
        Ranks Evaluated & \{1, 2, 4\} & \{1, 2, 4\} \\
        Dataset Size (Train) & 8000 & 8000 \\
        Dataset Size (Val) & 1000 & 1000 \\
        Batch Size & 256 & 256 \\
        \midrule
        \multicolumn{3}{c}{\textit{Pre-Training (Phase 1)}} \\
        \midrule
        Optimizer & Adam & Adam \\
        Learning Rate & $2 \times 10^{-3}$ & $2 \times 10^{-3}$ \\
        Sparsity Penalty ($\lambda_{L1}$) & $2 \times 10^{-4}$ & $2 \times 10^{-4}$ \\
        Epochs & 30 & 30 \\
        \midrule
        \multicolumn{3}{c}{\textit{Unlearning Attack (Phase 2)}} \\
        \midrule
        Optimizer & SGD & SGD \\
        Learning Rate & $2 \times 10^{-2}$ & $2 \times 10^{-2}$ \\
        Attack Steps & 50 & 50 \\
        Forget Loss Weight & 0.5 & 0.5 \\
        \bottomrule
    \end{tabular}
\end{table*}

\section{Extended Experimental Results}
\begin{figure}[t]
    \centering
    \includegraphics[width=\linewidth]{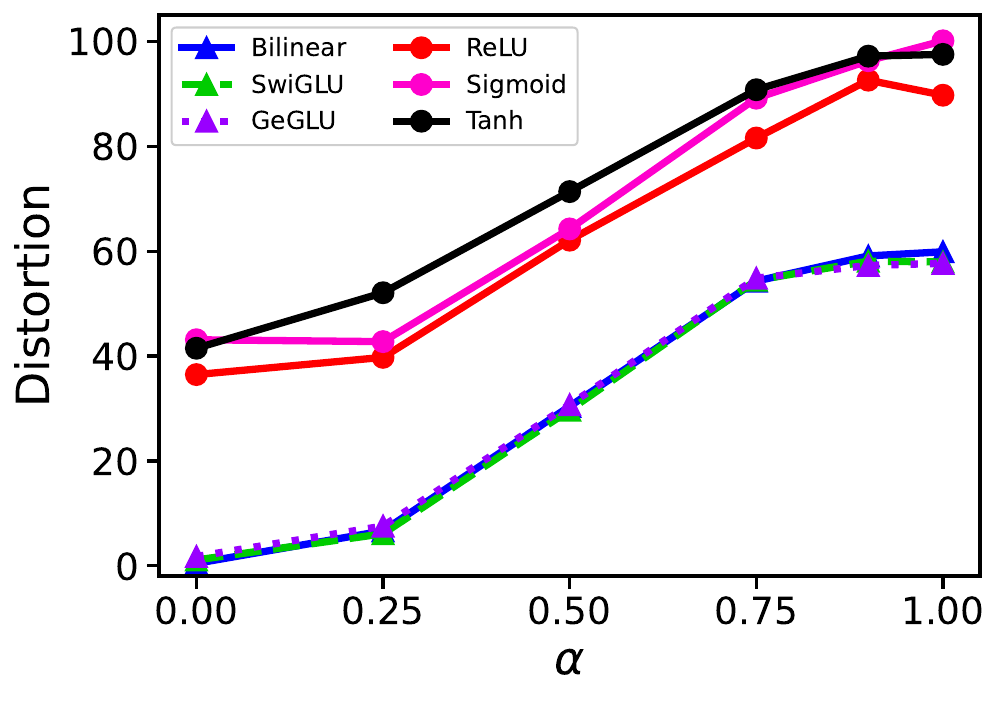}
    \vspace{-1em}
    \caption{\textbf{Task Preservation Distortion Across Architectures.} 
    Distortion measures the deviation from the expected Task B projection strength learned during Phase 1 training. 
    Multiplicative architectures exhibit near-zero distortion at $\alpha=0$ and maintain significantly lower distortion across all correlation levels. 
    Pointwise architectures show consistently high distortion ($\sim$40--100).}
    \label{fig:distortion_analysis}
    % \vspace{-1.3em}
\end{figure}

\begin{figure}[t]
    \centering
    \includegraphics[width=\columnwidth]{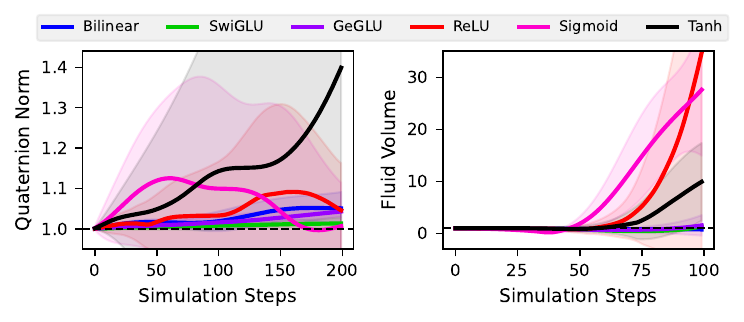}
    \caption{Lie group dynamics curves robustness results for 5 different seeds. \textbf{Left}: Rigid Body Stability (S3). \textbf{Right}: Fluid Volume Conservation (SL(2))}
    \label{fig:volume}
\end{figure}
\begin{figure*}[t]
    \centering
    \includegraphics[width=\linewidth]{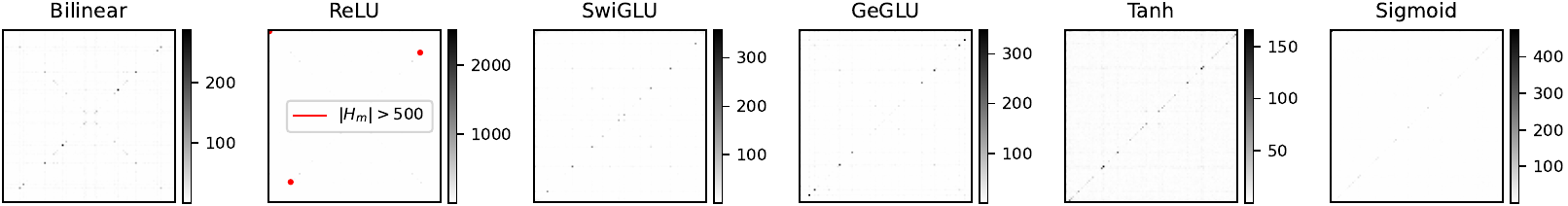}
    \caption{Fourier-domain magnitude of a representative interaction matrix $M_k$ for modular addition. The true operator exhibits a diagonal structure corresponding to the circulant addition kernel. Multiplicative architectures recover this diagonal Fourier geometry, whereas pointwise nonlinear baselines produce diffuse or highly sparse spectra, indicating misalignment with the underlying algebraic structure.
}
    \label{fig:add-heatmaps}
\end{figure*}

\label{app:extended_results}

% --- Helper command to center-align row labels ---
% 3.5em is approx half the height of the plots. Adjust this number if your plots are taller/shorter.

\subsection{Full Robustness Sweeps}

Here we visualize the unlearning dynamics as a continuous spectrum. Each row represents a specific architecture, and the columns represent the increasing task correlation $\alpha$ (from $0.0$ to $1.0$).

\begin{figure*}[h]
    \centering
    \makebox[\linewidth]{\textbf{Bilinear}}
    % --- ROW 1: BILINEAR ---
    \begin{subfigure}{0.155\linewidth} \vcenteredinclude{figures/bilinear_alpha_0.0.pdf} \end{subfigure}
    \hfill
    \begin{subfigure}{0.155\linewidth} \vcenteredinclude{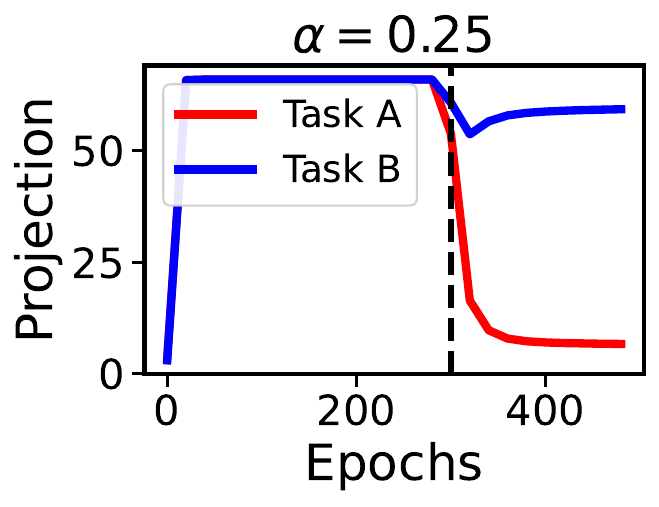} \end{subfigure}
    \hfill
    \begin{subfigure}{0.155\linewidth} \vcenteredinclude{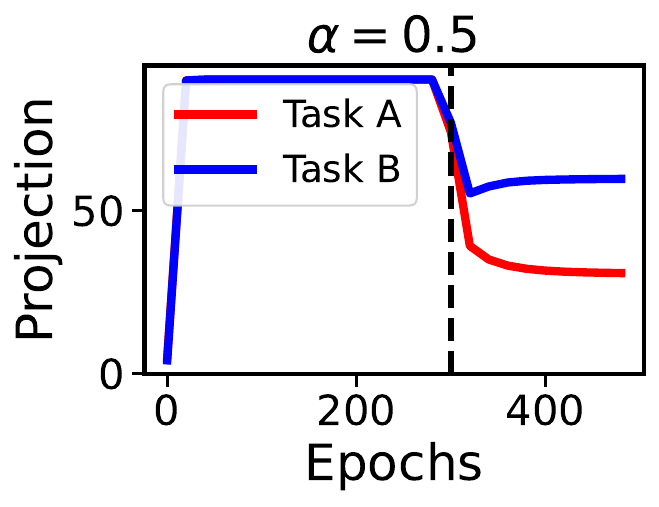} \end{subfigure}
    \hfill
    \begin{subfigure}{0.155\linewidth} \vcenteredinclude{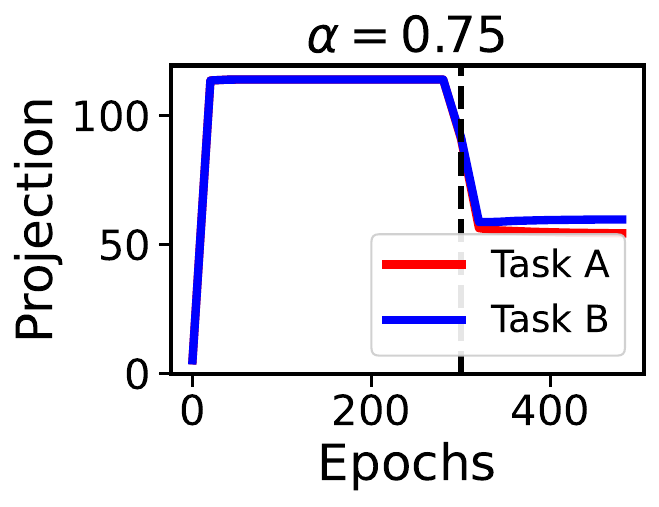} \end{subfigure}
    \hfill
    \begin{subfigure}{0.155\linewidth} \vcenteredinclude{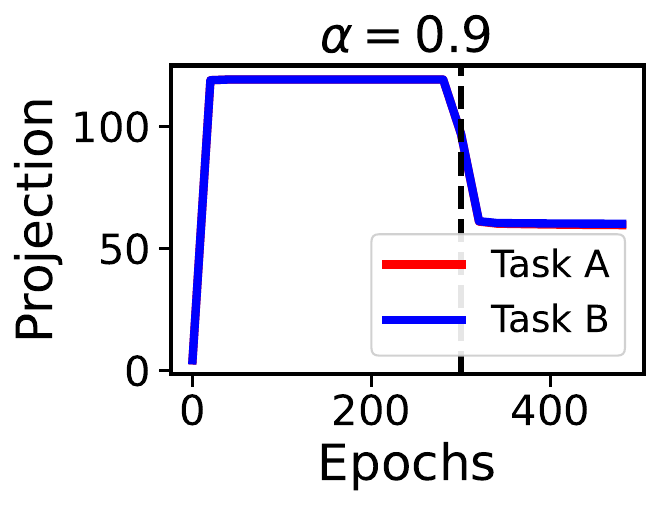} \end{subfigure}
    \hfill
    \begin{subfigure}{0.155\linewidth} \vcenteredinclude{figures/bilinear_alpha_1.0.pdf} \end{subfigure}
    
    \vspace{1em}
    \makebox[\linewidth]{\textbf{SwiGLU}}
    % --- ROW 2: SWIGLU ---
    \begin{subfigure}{0.155\linewidth} \vcenteredinclude{figures/swiglu_alpha_0.0.pdf} \end{subfigure}
    \hfill
    \begin{subfigure}{0.155\linewidth} \vcenteredinclude{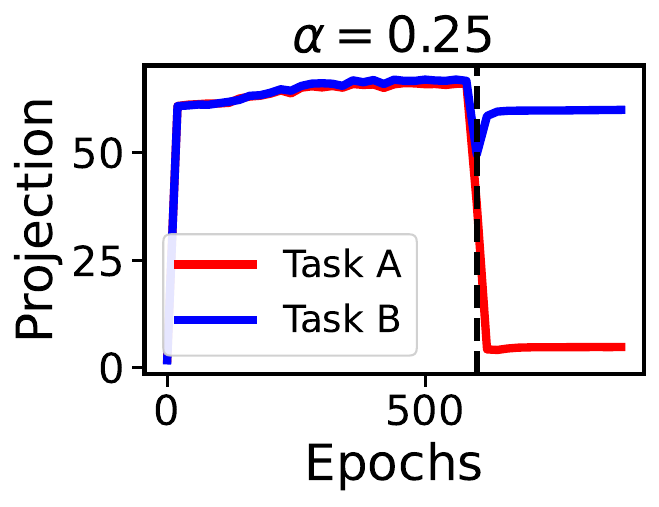} \end{subfigure}
    \hfill
    \begin{subfigure}{0.155\linewidth} \vcenteredinclude{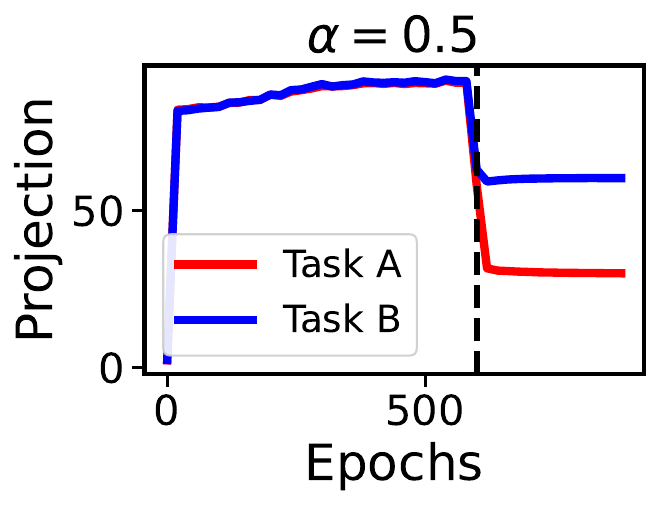} \end{subfigure}
    \hfill
    \begin{subfigure}{0.155\linewidth} \vcenteredinclude{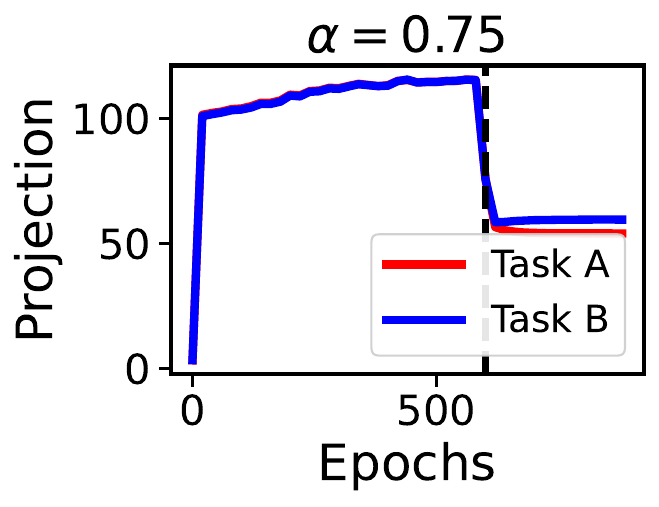} \end{subfigure}
    \hfill
    \begin{subfigure}{0.155\linewidth} \vcenteredinclude{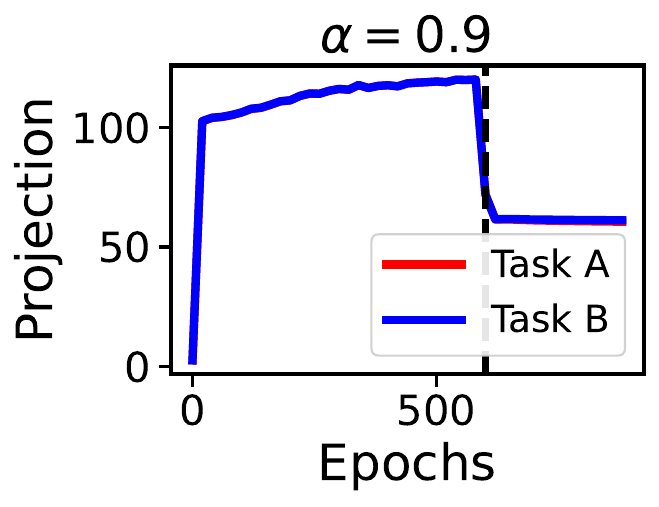} \end{subfigure}
    \hfill
    \begin{subfigure}{0.155\linewidth} \vcenteredinclude{figures/swiglu_alpha_1.0.pdf} \end{subfigure}

    \vspace{1em}
    \makebox[\linewidth]{\textbf{GeGLU}}
    % --- ROW 3: GEGLU ---
    \begin{subfigure}{0.155\linewidth} \vcenteredinclude{figures/geglu_alpha_0.0.pdf} \end{subfigure}
    \hfill
    \begin{subfigure}{0.155\linewidth} \vcenteredinclude{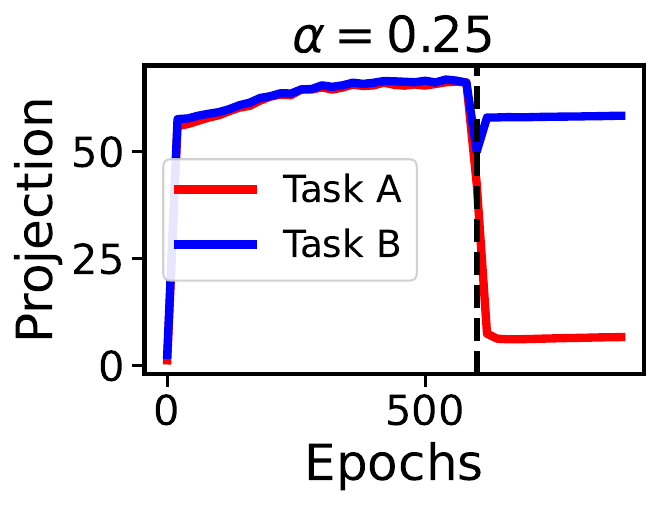} \end{subfigure}
    \hfill
    \begin{subfigure}{0.155\linewidth} \vcenteredinclude{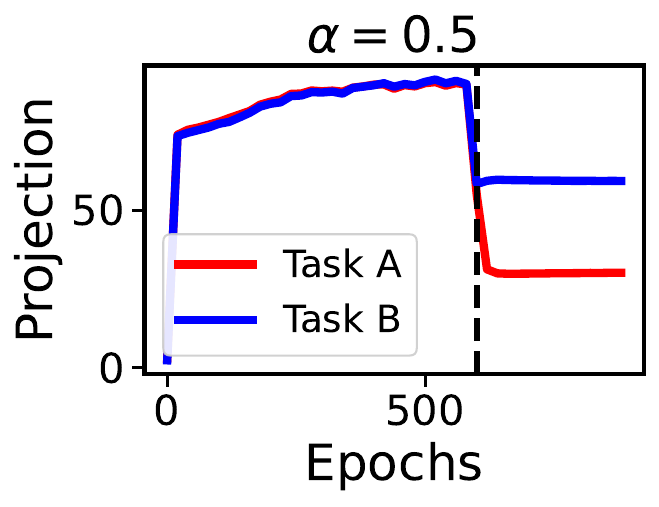} \end{subfigure}
    \hfill
    \begin{subfigure}{0.155\linewidth} \vcenteredinclude{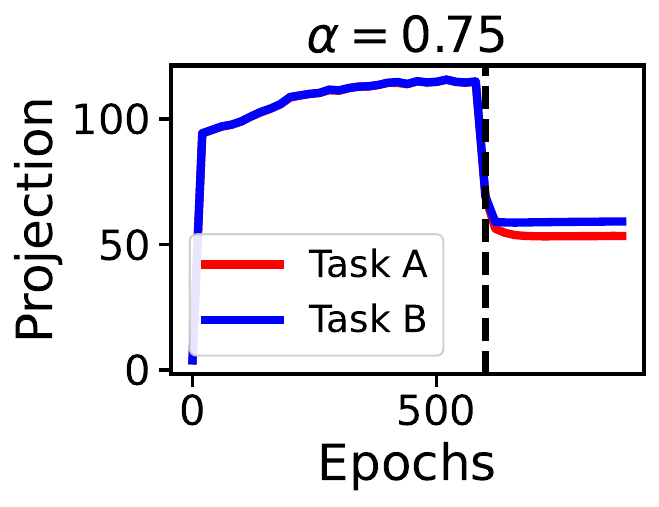} \end{subfigure}
    \hfill
    \begin{subfigure}{0.155\linewidth} \vcenteredinclude{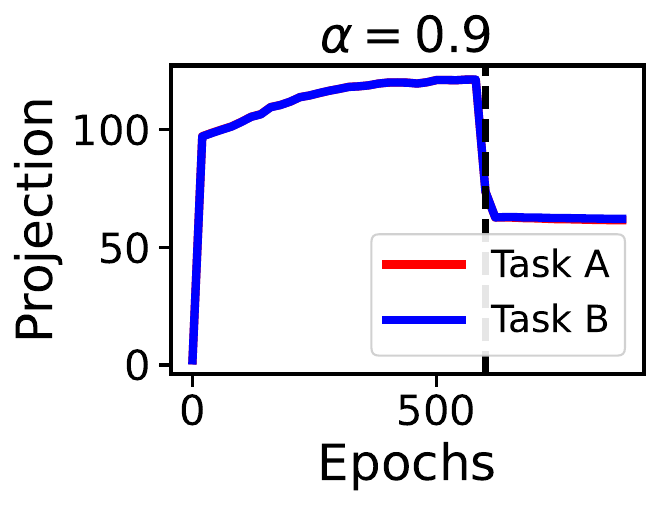} \end{subfigure}
    \hfill
    \begin{subfigure}{0.155\linewidth} \vcenteredinclude{figures/geglu_alpha_1.0.pdf} \end{subfigure}

    \caption{\textbf{Multiplicative Family robustness sweep.} Each row shows one architecture. Moving left to right, the task correlation $\alpha$ increases.}
    \label{fig:app_strip_multi}
\end{figure*}

\begin{figure*}[h]
    \centering
    \makebox[\linewidth]{\textbf{ReLU}}
    % --- ROW 1: RELU ---
    \begin{subfigure}{0.155\linewidth} \vcenteredinclude{figures/relu_alpha_0.0.pdf} \end{subfigure}
    \hfill
    \begin{subfigure}{0.155\linewidth} \vcenteredinclude{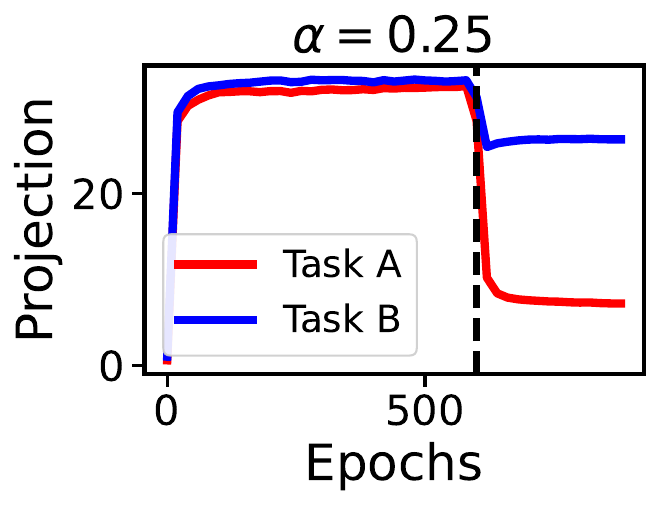} \end{subfigure}
    \hfill
    \begin{subfigure}{0.155\linewidth} \vcenteredinclude{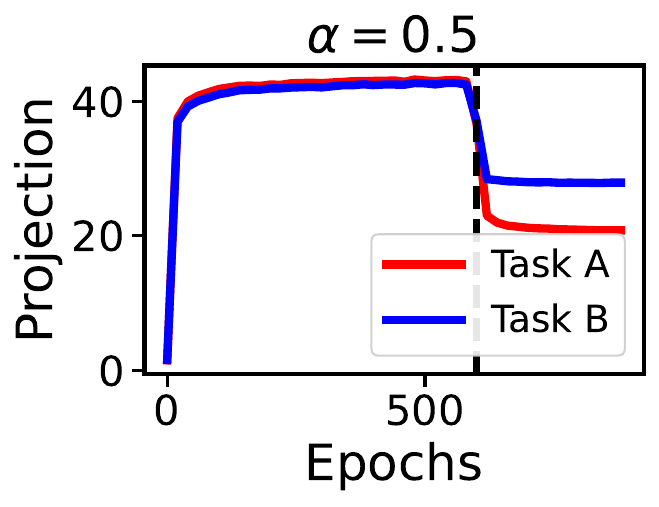} \end{subfigure}
    \hfill
    \begin{subfigure}{0.155\linewidth} \vcenteredinclude{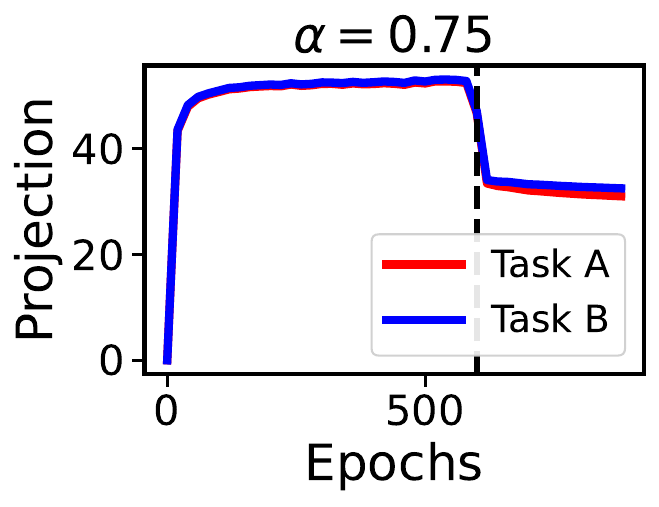} \end{subfigure}
    \hfill
    \begin{subfigure}{0.155\linewidth} \vcenteredinclude{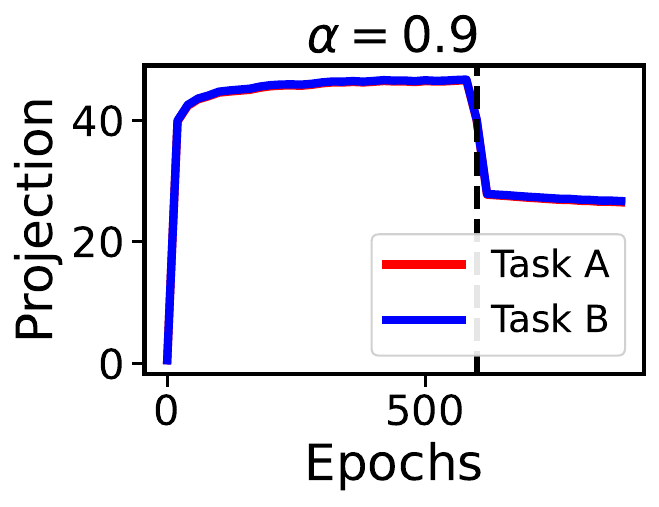} \end{subfigure}
    \hfill
    \begin{subfigure}{0.155\linewidth} \vcenteredinclude{figures/relu_alpha_1.0.pdf} \end{subfigure}
    
    \vspace{1em}
    \makebox[\linewidth]{\textbf{Tanh}}

    % --- ROW 2: TANH ---
    \begin{subfigure}{0.155\linewidth} \vcenteredinclude{figures/tanh_alpha_0.0.pdf} \end{subfigure}
    \hfill
    \begin{subfigure}{0.155\linewidth} \vcenteredinclude{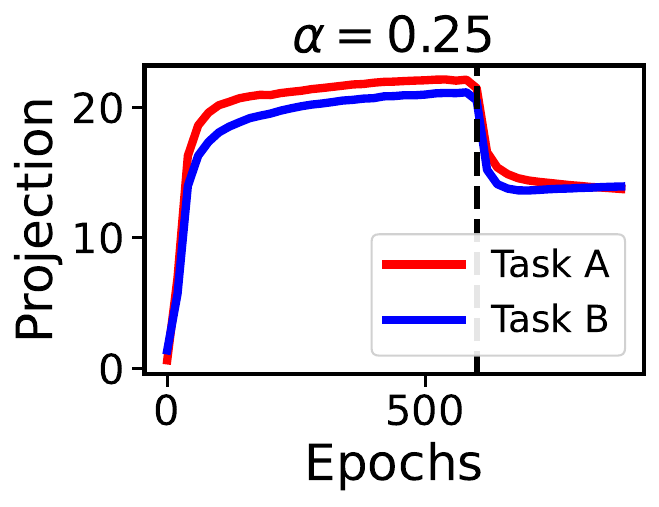} \end{subfigure}
    \hfill
    \begin{subfigure}{0.155\linewidth} \vcenteredinclude{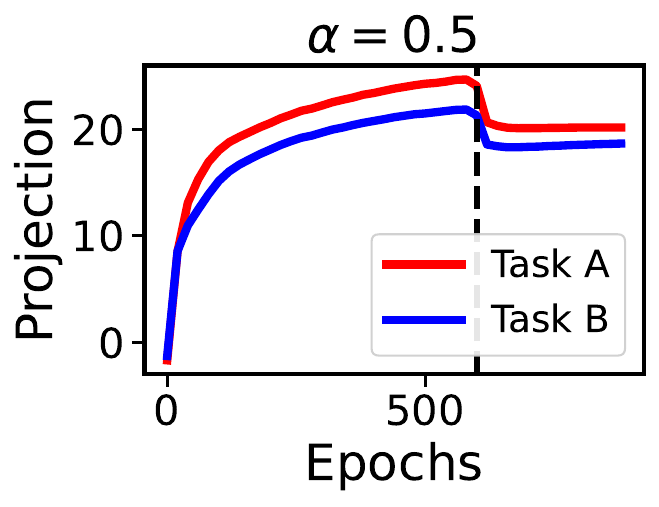} \end{subfigure}
    \hfill
    \begin{subfigure}{0.155\linewidth} \vcenteredinclude{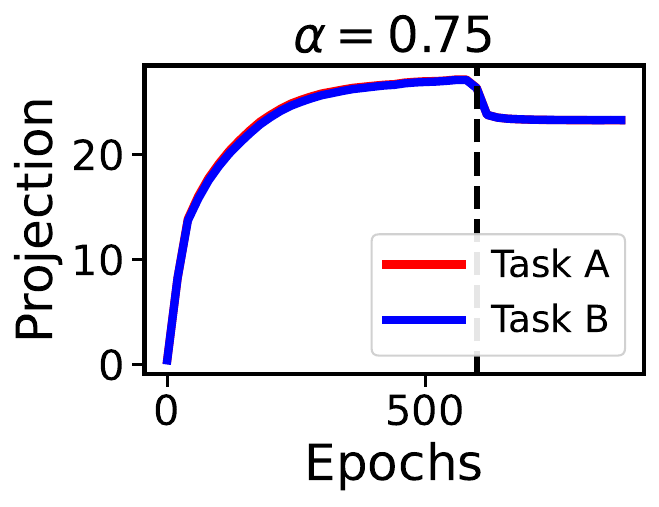} \end{subfigure}
    \hfill
    \begin{subfigure}{0.155\linewidth} \vcenteredinclude{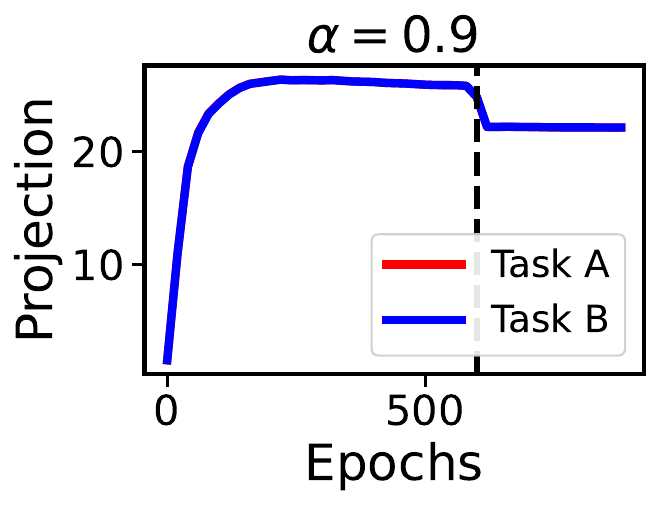} \end{subfigure}
    \hfill
    \begin{subfigure}{0.155\linewidth} \vcenteredinclude{figures/tanh_alpha_1.0.pdf} \end{subfigure}

    \vspace{1em}
    \makebox[\linewidth]{\textbf{Sigmoid}}

    % --- ROW 3: SIGMOID ---
    \begin{subfigure}{0.155\linewidth} \vcenteredinclude{figures/sigmoid_alpha_0.0.pdf} \end{subfigure}
    \hfill
    \begin{subfigure}{0.155\linewidth} \vcenteredinclude{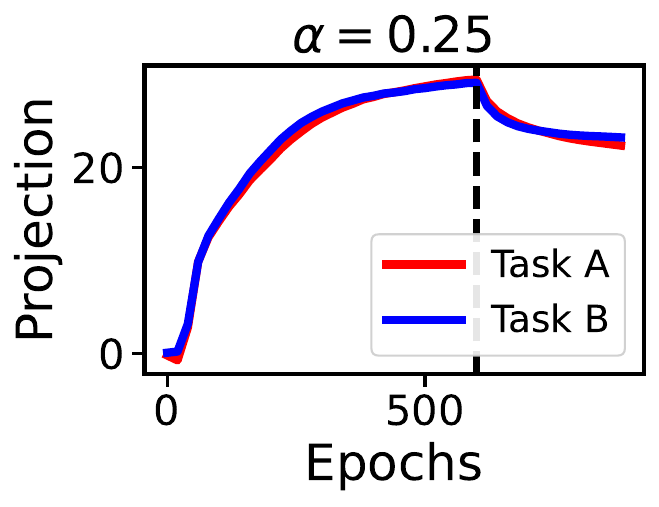} \end{subfigure}
    \hfill
    \begin{subfigure}{0.155\linewidth} \vcenteredinclude{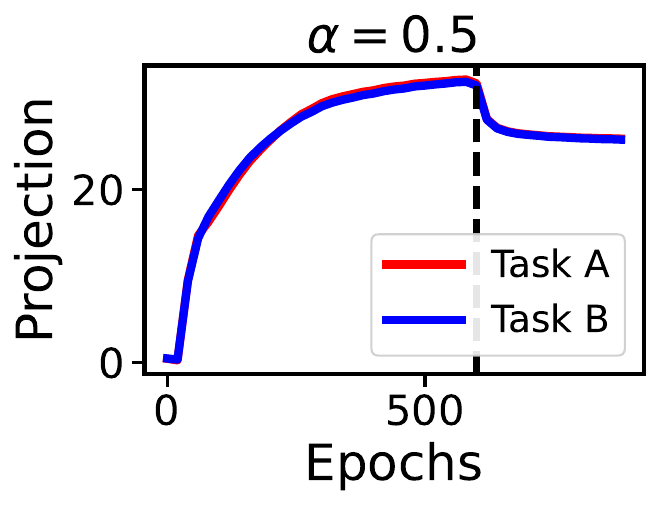} \end{subfigure}
    \hfill
    \begin{subfigure}{0.155\linewidth} \vcenteredinclude{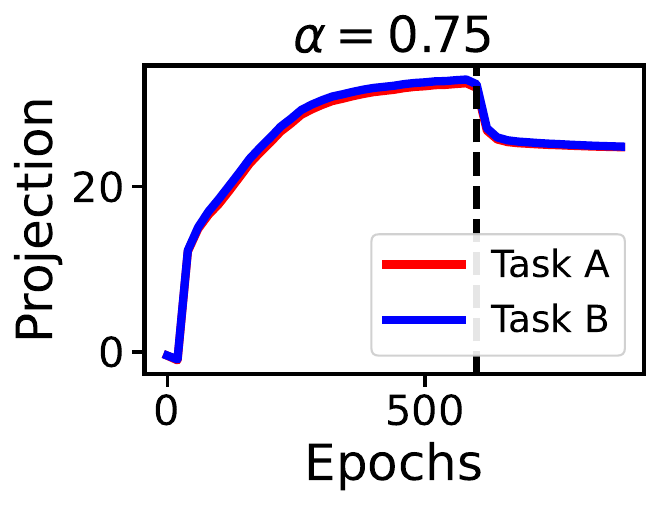} \end{subfigure}
    \hfill
    \begin{subfigure}{0.155\linewidth} \vcenteredinclude{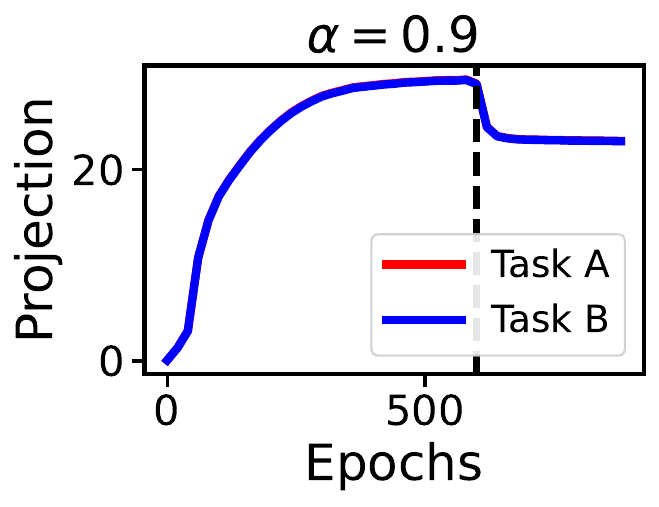} \end{subfigure}
    \hfill
    \begin{subfigure}{0.155\linewidth} \vcenteredinclude{figures/sigmoid_alpha_1.0.pdf} \end{subfigure}

    \caption{\textbf{Pointwise Family robustness sweep.} As correlation $\alpha$ increases (moving right), the unlearning performance degrades.}
    \label{fig:app_strip_point}
\end{figure*}

\begin{itemize}
    \item \textbf{Multiplicative Models (Fig.~\ref{fig:app_strip_multi}):} Note how the gap between the Blue line (Forgotten) and Red line (Retained) remains distinct even as you move to the far right columns.
    \item \textbf{Pointwise Models (Fig.~\ref{fig:app_strip_point}):} Note how the Red line collapses downwards as you move to the right, indicating interference.
\end{itemize}

\subsection{Modular Arithmetic on $\mathbb{Z}_{113}$}
\label{appendix:mod-113}

To verify that our observations are not specific to a particular modulus, we repeat the modular arithmetic experiments on the prime field $\mathbb{Z}_{113}$, using the same architectures, optimization settings, and evaluation metrics as in the main experiments on $\mathbb{Z}_{97}$. The only change is the modulus $p=113$, resulting in a larger input space and interaction table.

Across both modular addition and multiplication tasks, we observe qualitatively identical behavior. In modular addition, bilinear and gated multiplicative architectures recover interaction operators whose Fourier-domain structure remains concentrated and closely aligned with the circulant ground-truth operator. The resulting Fourier entropy values remain near the theoretical target of $\log p$, while pointwise nonlinear baselines either collapse to low-entropy procedural representations (ReLU, Sigmoid) or exhibit diffuse spectra (Tanh), consistent with the trends reported in the main text.

For modular multiplication, the learned interaction matrices continue to exhibit a sharp singular value decay under bilinear parameterizations, indicating effective low-rank factorization of the multiplication table. Pointwise architectures again display significantly slower spectral decay, suggesting entangled and distributed representations.

Overall, these results confirm that the structural disentanglement induced by multiplicative architectures is not sensitive to the choice of modulus and persists across different prime fields. This supports the interpretation that the observed effects arise from architectural inductive bias rather than properties specific to a particular problem instance.
\begin{figure}[ht!]
    \centering
    \includegraphics[width=\linewidth]{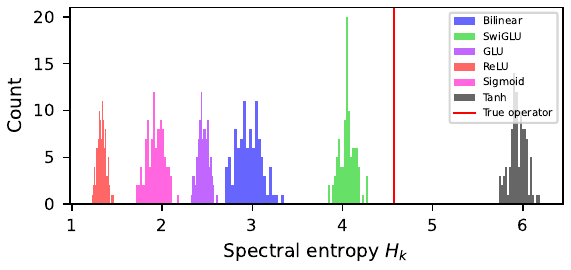}
    \caption{Spectral entropy $H_k$ for $p=113$.}
    \label{fig:add-entropy-113}
\end{figure}

\begin{figure}[t]
    \centering
    \includegraphics[width=\linewidth]{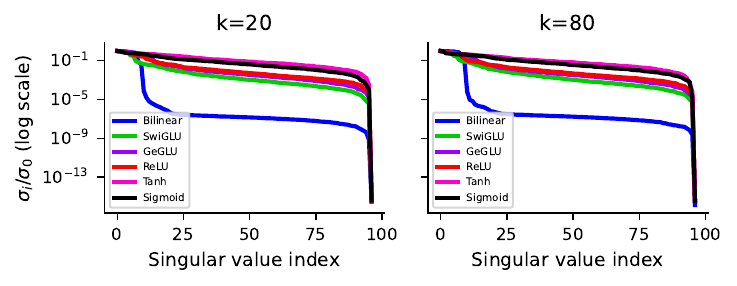}
    \caption{Normalized singular value decay for $p=113$.}
    \label{fig:mul-svd-113}
\end{figure}

\begin{figure*}[htbp]
    \centering
    \includegraphics[width=\linewidth]{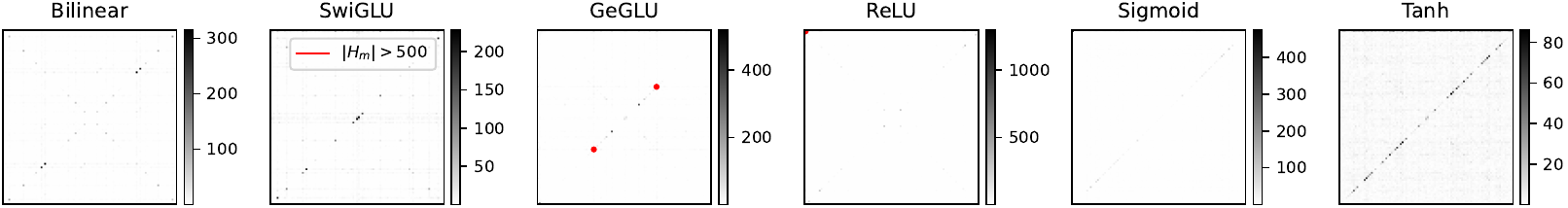}
    \caption{Fourier-domain magnitude for $p=113$.
}
    \label{fig:add-heatmaps-113}
\end{figure*}

\end{document}